\documentclass[11pt,twoside]{article}

\usepackage{amsfonts,epsfig,graphicx}
\usepackage{amsmath,amssymb,amsthm}
\usepackage{fullpage}

\usepackage{epsf}
\usepackage{fancyheadings}
\usepackage{graphics}
\usepackage{amsfonts}
\usepackage{amsmath}
\usepackage{psfrag}

\usepackage{color}

\usepackage[numbers]{natbib}

\theoremstyle{plain}

\newtheorem{theo}{Theorem}[section]

\newtheorem{lem}{Lemma}[section]
\newtheorem{prop}{Proposition}[section]
\newtheorem{cor}{Corollary}[section]

\theoremstyle{definition} 

\newtheorem{nota}{Notation}[section]
\newtheorem{de}{Definition}[section]
\newtheorem{exa}{Example}[section]
\newtheorem{as}{Assumption}[section]
\newtheorem{alg}{Algorithm}[section]

\newcommand{\btheo}{\begin{theo}}
\newcommand{\bde}{\begin{de}}
\newcommand{\ble}{\begin{lem}}
\newcommand{\bpr}{\begin{prop}}
\newcommand{\bno}{\begin{nota}}
\newcommand{\bex}{\begin{exa}}
\newcommand{\bcor}{\begin{cor}}
\newcommand{\spro}{\begin{proof}}
\newcommand{\bas}{\begin{as}}
\newcommand{\balg}{\begin{alg}}

\newcommand{\etheo}{\end{theo}}
\newcommand{\ede}{\end{de}}
\newcommand{\ele}{\end{lem}}
\newcommand{\epr}{\end{prop}}
\newcommand{\eno}{\end{nota}}
\newcommand{\eex}{\end{exa}}
\newcommand{\ecor}{\end{cor}}
\newcommand{\fpro}{\end{proof}}
\newcommand{\eas}{\end{as}}
\newcommand{\ealg}{\end{alg}}

\theoremstyle{plain}

\newtheorem{theos}{Theorem}
\newtheorem{props}{Proposition}
\newtheorem{lems}{Lemma}
\newtheorem{cors}{Corollary}

\theoremstyle{definition}
\newtheorem{exas}{Example}
\newtheorem{algs}{Algorithm}
\newtheorem{asss}{Assumption}
\newtheorem{defns}{Definition}

\newcommand{\btheos}{\begin{theos}}
\newcommand{\etheos}{\end{theos}}
\newcommand{\bprops}{\begin{props}}
\newcommand{\eprops}{\end{props}}
\newcommand{\bdes}{\begin{defns}}
\newcommand{\edes}{\end{defns}}
\newcommand{\blems}{\begin{lems}}
\newcommand{\elems}{\end{lems}}
\newcommand{\bcors}{\begin{cors}}
\newcommand{\ecors}{\end{cors}}
\newcommand{\bexs}{\begin{exas}}
\newcommand{\eexs}{\end{exas}}
\newcommand{\balgs}{\begin{algs}}
\newcommand{\ealgs}{\end{algs}}
\newcommand{\bass}{\begin{asss}}
\newcommand{\eass}{\end{asss}}

\newcommand{\numobs}{\ensuremath{n}}
\newcommand{\usedim}{\ensuremath{d}}

\newcommand{\plaincon}{\ensuremath{c}}

\newcommand{\mprob}{\ensuremath{\mathbb{P}}}

\newcommand{\numproj}{\ensuremath{m}}

\newcommand{\Sketch}{\ensuremath{S}}

\newcommand{\yvec}{\ensuremath{y}}

\newcommand{\real}{\ensuremath{\mathbb{R}}}
\newcommand{\defn}{\ensuremath{: \, =}}
\newcommand{\inprod}[2]{\ensuremath{\langle #1 , \, #2 \rangle}}

\newcommand{\KerMat}{\ensuremath{K}}

\newcommand{\Xspace}{\ensuremath{\mathcal{X}}}
\newcommand{\fstar}{\ensuremath{f^*}}

\newcommand{\alphahat}{\ensuremath{\widehat{\alpha}}}

\newcommand{\Sphere}[1]{\ensuremath{\mathcal{S}^{#1-1}}}
\newcommand{\SPHERE}[1]{\ensuremath{\Sphere{#1}}}

\newcommand{\fhat}{\ensuremath{\widehat{f}}}
\newcommand{\ftil}{\ensuremath{\widetilde{f}}}
\newcommand{\KerPlain}{\ensuremath{\mathcal{K}}}

\newcommand{\KerFun}[2]{\ensuremath{\KerPlain(#1, #2)}}

\newcommand{\betastar}{\ensuremath{\beta^*}}

\newcommand{\Term}{\ensuremath{T}}

\newcommand{\CEXP}[1]{\ensuremath{e^{#1}}}

\newcommand{\Hil}{\ensuremath{\mathcal{H}}}

\newcommand{\order}{\ensuremath{\mathcal{O}}}

\newcommand{\Ellipse}{\ensuremath{\mathcal{E}}}

\newcommand{\myspan}{\ensuremath{\operatorname{span}}}

\newcommand{\diag}{\ensuremath{\operatorname{diag}}}

\newcommand{\PlainKerFun}{\ensuremath{\mathcal{K}}}

\newlength{\widebarargwidth}
\newlength{\widebarargheight}
\newlength{\widebarargdepth}
\DeclareRobustCommand{\widebar}[1]{%
  \settowidth{\widebarargwidth}{\ensuremath{#1}}%
  \settoheight{\widebarargheight}{\ensuremath{#1}}%
  \settodepth{\widebarargdepth}{\ensuremath{#1}}%
  \addtolength{\widebarargwidth}{-0.3\widebarargheight}%
  \addtolength{\widebarargwidth}{-0.3\widebarargdepth}%
  \makebox[0pt][l]{\hspace{0.3\widebarargheight}%
    \hspace{0.3\widebarargdepth}%
    \addtolength{\widebarargheight}{0.3ex}%
    \rule[\widebarargheight]{0.95\widebarargwidth}{0.1ex}}%
  {#1}}

\newcommand{\kull}[2]{\ensuremath{D(#1\; \| \; #2)}}

\newcommand{\Event}{\ensuremath{\mathcal{E}}}
\newcommand{\ytil}{\ensuremath{\widetilde{\yvec}}}

\newcommand{\widgraph}[2]{\includegraphics[keepaspectratio,width=#1]{#2}}

\newcommand{\alphadagger}{\ensuremath{\alpha^\dagger}}
\newcommand{\fdagger}{\ensuremath{f^\dagger}}

\DeclareMathOperator{\trace}{trace}

\newcommand{\UNICON}{\ensuremath{c}}

\newcommand{\DelHat}{\ensuremath{\widehat{\Delta}}}
\newcommand{\DelTil}{\ensuremath{\widetilde{\Delta}}}

\newcommand{\delcrit}{\ensuremath{\delta_\numobs}}
\newcommand{\KerComp}{\ensuremath{\mathcal{R}}}
\newcommand{\effdim}{\statdim}

\newcommand{\thetastar}{\ensuremath{\theta^*}}
\newcommand{\thetatil}{\ensuremath{\widetilde{\theta}}}

\newcommand{\statdim}{\ensuremath{{\usedim_\numobs}}}

\newcommand{\SketchTil}{\ensuremath{\widetilde{\Sketch}}}

\newcommand{\matsnorm}[2]{|\!|\!| #1 | \! | \!|_{{#2}}}

\newcommand{\opnorm}[1]{\ensuremath{\matsnorm{#1}{\tiny{\mbox{op}}}}}

\newcommand{\Ztil}{\ensuremath{\widetilde{Z}}}

\newcommand{\regparn}{\ensuremath{\lambda_\numobs}}
\newcommand{\regpar}{\ensuremath{\lambda}}

\newcommand{\FstarEval}{\ensuremath{z^*}}

\newcommand{\PRECON}{\ensuremath{c}} 

\newcommand{\empkereig}{\ensuremath{\widehat{\mu}}} 
\newcommand{\EmpKerComp}{\ensuremath{\widehat{\KerComp}}}

\newcommand{\popkereig}{\ensuremath{\mu}} 
\newcommand{\PopKerComp}{\ensuremath{\KerComp}}

\newcommand{\clower}{c_\ell}
\newcommand{\cupper}{c_u}

\newcommand{\flest}{\ensuremath{\ftil}}

\newcommand{\MYHACKCON}{\ensuremath{c}}

\newcommand{\hackmu}[1]{\ensuremath{ \min \{\delta^2, \empkereig_{#1}
    \}}}
\newcommand{\ELLNORM}[1]{\ensuremath{\|#1\|_{\Ellipse}}}

\newcommand{\weight}{\ensuremath{\omega}}
\newcommand{\wdagger}{\ensuremath{\weight^\dagger}}

\newcommand{\BasicEllipse}{\ensuremath{\mathcal{B}}}

\newcommand{\RandInd}{\ensuremath{A}}
\newcommand{\randind}{\ensuremath{a}}
\newcommand{\randindb}{\ensuremath{b}}
\newcommand{\radevec}{\ensuremath{r}}

\newcommand{\vtil}{\ensuremath{\widetilde{v}}}
\newcommand{\Qtil}{\ensuremath{\widetilde{Q}}}
\newcommand{\Dbar}{\ensuremath{\widebar{D}}}

\newcommand{\Exs}{\ensuremath{\mathbb{E}}}

\newcommand{\algrank}{\ensuremath{r}}

\newcommand{\ptil}{\ensuremath{\tilde{p}}}

\newcommand{\PlainKerFunLip}{\ensuremath{\PlainKerFun_{\mbox{\scriptsize{sob}}}}}
\newcommand{\KerFunLip}[2]{\ensuremath{\PlainKerFunLip(#1, #2)}}

\newcommand{\PlainKerFunGauss}{\ensuremath{\PlainKerFun_{\mbox{\scriptsize{gau}}}}}
\newcommand{\KerFunGauss}[2]{\ensuremath{\PlainKerFunGauss(#1, #2)}}

\newcommand{\PlainKerFunPoly}{\ensuremath{\PlainKerFun_{\mbox{\scriptsize{poly}}}}}
\newcommand{\KerFunPoly}[2]{\ensuremath{\PlainKerFunPoly(#1, #2)}}

\newcommand{\HACK}{2}
\newcommand{\TWOHACK}{\frac{1}{16}}

\newcommand{\PHIPROB}{\phi(\numproj, \statdim, \numobs)}

\newcommand{\RADE}{\ensuremath{R}}

\newcommand{\polydeg}{\ensuremath{D}}
\newcommand{\bcar}{\begin{carrlist}}
\newcommand{\ecar}{\end{carlist}}

\newcommand{\NYS}{Nystr\"{o}m~}

\newcommand{\pcol}{\ensuremath{p}} 

\long\def\comment#1{}

\makeatletter
\long\def\@makecaption#1#2{
        \vskip 0.8ex
        \setbox\@tempboxa\hbox{\small {\bf #1:} #2}
        \parindent 1.5em  %% How can we use the global value of this???
        \dimen0=\hsize
        \advance\dimen0 by -3em
        \ifdim \wd\@tempboxa >\dimen0
                \hbox to \hsize{
                        \parindent 0em
                        \hfil 
                        \parbox{\dimen0}{\def\baselinestretch{0.96}\small
                                {\bf #1.} #2
                                } 
                        \hfil}
        \else \hbox to \hsize{\hfil \box\@tempboxa \hfil}
        \fi
        }
\makeatother

\begin{document}

\begin{center} {\LARGE{\bf{Randomized sketches for kernels: \\
Fast and optimal non-parametric regression}}} \\
  \vspace{1cm}

\begin{center}
\begin{tabular}{ccccc}
Yun Yang$^1$ && Mert Pilanci$^1$ && Martin J. Wainwright$^{1,2}$ 
\end{tabular}
\end{center}

  \vspace{.5cm}
  {\large University of California, Berkeley} \\
  \vspace{.15cm} $^1$Department of Electrical Engineering and Computer
  Science ~~~~ $^2$Department of Statistics

\vspace*{.2in}

\today

\end{center}

\vspace*{.5in}

%%%%%%%%%%%%%%%%%%%%%%%%%%%%%%%%%%%%%%%%%%%%%%%%%%%%%%%%%%%%%%%%%%%%%%%

\begin{abstract}
Kernel ridge regression (KRR) is a standard method for performing
non-parametric regression over reproducing kernel Hilbert spaces.
Given $n$ samples, the time and space complexity of computing the KRR
estimate scale as $\mathcal{O}(n^3)$ and $\mathcal{O}(n^2)$
respectively, and so is prohibitive in many cases.  We propose
approximations of KRR based on $m$-dimensional randomized sketches of
the kernel matrix, and study how small the projection dimension $m$
can be chosen while still preserving minimax optimality of the
approximate KRR estimate.  For various classes of randomized sketches,
including those based on Gaussian and randomized Hadamard matrices, we
prove that it suffices to choose the sketch dimension $m$ proportional
to the statistical dimension (modulo logarithmic factors).  Thus, we
obtain fast and minimax optimal approximations to the KRR estimate for
non-parametric regression.
\end{abstract}

%%%%%%%%%%%%%%%%%%%%%%%%%%%%%%%%%%%%%%%%%%%%%%%%%%%%%%%%%%%%%%%%%%%%%%%%%

\section{Introduction}

The goal of non-parametric regression is to make predictions of a
response variable $Y \in \real$ based on observing a covariate vector
$X \in \Xspace$.  In practice, we are given a collection of $\numobs$
samples, say $\{(x_i, y_i)\}_{i=1}^\numobs$ of covariate-response
pairs and our goal is to estimate the regression function $f^*(x) =
\Exs[Y \mid X = x]$.  In the standard Gaussian model, it is assumed
that the covariate-response pairs are related via the model
\begin{align}
\label{EqnModel}
y_i & = \fstar(x_i) + \sigma w_i, \quad \mbox{for $i = 1, \ldots,
  \numobs$}
\end{align}
where the sequence $\{w_i\}_{i=1}^\numobs$ consists of i.i.d.\!
standard Gaussian variates.  It is typical to assume that the
regression function $\fstar$ has some regularity properties, and one
way of enforcing such structure is to require $f^*$ to belong to a
reproducing kernel Hilbert space, or RKHS for
short~\cite{Aronszajn50,Wahba,Gu02}).  Given such an assumption, it is
natural to estimate $\fstar$ by minimizing a combination of the
least-squares fit to the data and a penalty term involving the squared
Hilbert norm, leading to an estimator known \emph{kernel ridge
  regression}, or KRR for short~\cite{Hastie01,Shawe04}).  From a
statistical point of view, the behavior of KRR can be characterized
using existing results on $M$-estimation and empirical processes
(e.g.~\cite{Koltchinskii06,Mendelson02,vandeGeer}).  When the
regularization parameter is set appropriately, it is known to yield a
function estimate with minimax prediction error for various classes of
kernels.

Despite these attractive statistical properties, the computational
complexity of computing the KRR estimate prevent it from being
routinely used in large-scale problems.  More precisely, in a standard
implementation~\cite{Saunders1998}, the time complexity and space
complexity of KRR in a standard implementation scale as
$\order(\numobs^3)$ and $\order(\numobs^2)$, respectively, where
$\numobs$ refers to the number of samples.  As a consequence, it
becomes important to design methods to compute approximate forms of
the KRR estimate, while retaining guarantees of optimality in terms of
statistical minimaxity.  Various authors have taken different approaches
to this problem.  Zhang et al.~\cite{ZhaDucWai_COLT13} analyze a
distributed implementation of KRR, in which a set of $t$ machines each
compute a separate estimate based on a random $t$-way partition of the
full data set, and combine it into a global estimate by averaging.
This divide-and-conquer approach has time complexity and space
complexity $\order(\numobs^3/t^3)$ and $\order (\numobs^2/t^2)$,
respectively. Zhang et al. \cite{ZhaDucWai_COLT13} give conditions on
the number of splits $t$, as a function of the kernel, under which
minimax optimality of the resulting estimator can be guaranteed.  More
closely related to this paper are methods that are based on forming a
low-rank approximation to the $\numobs$-dimensional kernel matrix,
such as the Nystr\"{o}m methods (e.g. \cite{Drineas05,Gittens13}).
The time complexity by using a low-rank approximation is either
$\order(\numobs \algrank^2)$ or $\order(\numobs^2 \algrank)$,
depending on the specific approach (excluding the time for factorization), where $\algrank$ is the maintained
rank, and the space complexity is $\order(\numobs \algrank)$. Some
recent work~\cite{Bach12,Alaoui14} analyzes the tradeoff between the
rank $\algrank$ and the resulting statistical performance of the
estimator, and we discuss this line of work at more length in
Section~\ref{SecNystrom}.

In this paper, we consider approximations to KRR based on random
projections, also known as sketches, of the data.  Random projections
are a classical way of performing dimensionality reduction, and are
widely used in many algorithmic contexts (e.g., see the
book~\cite{Vem04} and references therein).  Our proposal is to
approximate $\numobs$-dimensional kernel matrix by projecting its row
and column subspaces to a randomly chosen $\numproj$-dimensional
subspace with $\numproj \ll \numobs$.  By doing so, an approximate
form of the KRR estimate can be obtained by solving an
$\numproj$-dimensional quadratic program, which involves time and
space complexity $\order(\numproj^3)$ and $\order(\numproj^2)$.
Computing the approximate kernel matrix is a pre-processing step that
has time complexity $\order(\numobs^2 \log(\numproj))$ for suitably
chosen projections; this pre-processing step is trivially
parallelizable, meaning it can be reduced to to $\order(\numobs^2
\log(\numproj)/t)$ by using $t \leq \numobs$ clusters.

Given such an approximation, we pose the following question: how small
can the projection dimension $\numproj$ be chosen while still
retaining minimax optimality of the approximate KRR estimate?  We
answer this question by connecting it to the \emph{statistical
  dimension} $\statdim$ of the $\numobs$-dimensional kernel matrix, a
quantity that measures the effective number of degrees of
freedom. (See Section~\ref{SecKernelComplexity} for a precise
definition.) From the results of earlier work on random projections for constrained Least Squares estimators (e.g., see \cite{PilWai14a,PilWai14b}), it is natural to conjecture that it should be possible
to project the kernel matrix down to the statistical dimension while
preserving minimax optimality of the resulting estimator.  The main
contribution of this paper is to confirm this conjecture for several
classes of random projection matrices.

The remainder of this paper is organized as follows.
Section~\ref{SecBackground} is devoted to further background on
non-parametric regression, reproducing kernel Hilbert spaces and
associated measures of complexity, as well as the notion of
statistical dimension of a kernel.  In Section~\ref{SecMain}, we turn
to statements of our main results.  Theorem~\ref{ThmMain0} provides a
general sufficient condition on a random sketch for the associated
approximate form of KRR to achieve the minimax risk.  In
Corollary~\ref{CorSubgaussian}, we derive some consequences of this
general result for particular classes of random sketch matrices, and
confirm these theoretical predictions with some simulations.  We also
compare at more length to methods based on the \NYS approximation in
Section~\ref{SecNystrom}.  Section~\ref{SecProofs} is devoted to the
proofs of our main results, with the proofs of more technical results
deferred to the appendices.  We conclude with a discussion in
Section~\ref{SecDiscussion}.

%%%%%%%%%%%%%%%%%%%%%%%%%%%%%%%%%%%%%%%%%%%%%%%%%%%%%%%%%%%%%%%%

\section{Problem formulation and background}
\label{SecBackground}

We begin by introducing some background on nonparametric regression
and reproducing kernel Hilbert spaces, before formulating the problem
discussed in this paper.

\subsection{Regression in reproducing kernel Hilbert spaces}
\label{SectionRK}

Given $\numobs$ samples $\{(x_i, y_i) \}_{i=1}^\numobs$ from the
non-parametric regression model~\eqref{EqnModel}, our goal is to
estimate the unknown regression function $\fstar$.  The quality of an
estimate $\fhat$ can be measured in different ways: in this paper, we
focus on the squared $L^2(\mprob_\numobs)$ error
\begin{align}
\label{EqnPredError}
\|\fhat - \fstar\|_\numobs^2 & \defn \frac{1}{\numobs}
\sum_{i=1}^\numobs \big(\fhat(x_i) - \fstar(x_i) \big)^2.
\end{align}
Naturally, the difficulty of non-parametric regression is controlled
by the structure in the function $\fstar$, and one way of modeling
such structure is within the framework of a reproducing kernel Hilbert
space (or RKHS for short).  Here we provide a very brief introduction
referring the reader to the books~\cite{BerTho04,Gu02,Wahba} for more
details and background.

Given a space $\Xspace$ endowed with a probability distribution
$\mprob$, the space $L^2(\mprob)$ consists of all functions that are
square-integrable with respect to $\mprob$.  In abstract terms, a
space $\Hil \subset L^2(\mprob)$ is an RKHS if for each $x
\in \Xspace$, the evaluation function $f \mapsto f(x)$ is a bounded
linear functional.  In more concrete terms, any RKHS is generated by a
positive semidefinite (PSD) kernel function in the following way.  A
PSD kernel function is a symmetric function $\PlainKerFun: \Xspace
\times \Xspace \rightarrow \real$ such that, for any positive integer
$N$, collections of points $\{v_1, \ldots, v_N\}$ and weight vector
$\weight \in \real^N$, the sum $\sum_{i,j=1}^N \weight_i \weight_j
\KerFun{v_i}{v_j}$ is non-negative.  Suppose moreover that for each
fixed $v \in \Xspace$, the function $u \mapsto \KerFun{u}{v}$ belongs
to $L^2(\mprob)$.  We can then consider the vector space of all
functions $g: \Xspace \rightarrow \real$ of the form
\begin{align*}
&\qquad g(\cdot)  = \sum_{i=1}^N \omega_i \KerFun{\cdot}{v_i}
\end{align*}
for some integer $N$, points $\{v_1, \ldots, v_N \} \subset \Xspace$ and
weight vector $w \in \real^N$.
By taking the closure of all such linear combinations, it can be
shown~\cite{Aronszajn50} that we generate an RKHS, and one that is
uniquely associated with the kernel $\PlainKerFun$.  We provide some
examples of various kernels and the associated function classes in
Section~\ref{SecKernelComplexity} to follow.

%%%%%%%%%%%%%%%%%%%%%%%%%%%%%%%%%%%%%%%%%%%%%%%%%%%%%%%%%%%%%%%%%%%%%%%%%

\subsection{Kernel ridge regression and its sketched form}
\label{SecKernelSketchBackground}
Given the dataset $\{(x_i,y_i)\}_{i=1}^n$, a natural method for
estimating unknown function $\fstar\in\Hil$ is known as kernel ridge
regression (KRR): it is based on the convex program
\begin{align}
\label{EqnKRR}
\fdagger & \defn \arg \min_{f \in \Hil} \Big\{\frac{1}{2 \numobs}
\sum_{i=1}^\numobs \big(y_i - f(x_i) \big)^2 + \regparn
\|f\|_{\Hil}^2\Big\},
\end{align}
where $\regparn$ is a regularization parameter.  

As stated, this optimization problem can be infinite-dimensional in
nature, since it takes place over the Hilbert space.  However, as a
straightforward consequence of the representer
theorem~\cite{Kimeldorf71}, the solution to this optimization problem
can be obtained by solving the $\numobs$-dimensional convex program.
In particular, let us define the \emph{empirical kernel matrix},
namely the $\numobs$-dimensional symmetric matrix $\KerMat$ with
entries $\KerMat_{ij} = \numobs^{-1} \KerFun{x_i}{x_j}$.  Here we
adopt the $\numobs^{-1}$ scaling for later theoretical convenience.
In terms of this matrix, the KRR estimate can be obtained by first
solving the quadratic program
\begin{subequations}
\begin{align}
\label{EqnOriginalKRR}
\wdagger & = \arg \min_{\weight \in \real^\numobs} \Big \{ \frac{1}{2}
\weight^T \KerMat^2 \weight - \weight^T \frac{\KerMat
  y}{\sqrt{\numobs}} + \regparn \weight^T \KerMat \weight \Big \},
\end{align}
and then outputting the function 
\begin{align}
\label{EqnOriginalKRREstimate}
\fdagger(\cdot) = \frac{1}{\sqrt{\numobs}} \sum_{i=1}^\numobs
\wdagger_i \KerFun{\cdot}{x_i}.
\end{align}
\end{subequations}

In principle, the original KRR optimization
problem~\eqref{EqnOriginalKRR} is simple to solve: it is an $\numobs$
dimensional quadratic program, and can be solved exactly using
$\order(\numobs^3)$ via a QR decomposition.  However, in many
applications, the number of samples may be large, so that this type of
cubic scaling is prohibitive.  In addition, the $\numobs$-dimensional
kernel matrix $\KerMat$ is dense in general, and so requires storage
of order $\numobs^2$ numbers, which can also be problematic in
practice.

In this paper, we consider an approximation based on limiting the
original parameter $\weight \in \real^\numobs$ to an
$\numproj$-dimensional subspace of $\real^\numobs$, where $\numproj
\ll \numobs$ is the \emph{projection dimension}.  We define this
approximation via a sketch matrix $\Sketch \in \real^{\numproj \times
  \numobs}$, such that the $\numproj$-dimensional subspace is
generated by the row span of $\Sketch$.  More precisely, the
\emph{sketched kernel ridge regression} estimate is given by first
solving
\begin{subequations}
\begin{align}
\label{EqnSketchedKRR}
\alphahat & = \arg \min_{\theta \in \real^\numproj} \Big \{
\frac{1}{2} \alpha^T (\Sketch \KerMat) (\KerMat \Sketch^T) \alpha -
\alpha^T\Sketch\, \frac{\KerMat \yvec}{\sqrt{\numobs}} +
\lambda_\numobs \alpha^T \Sketch \KerMat \Sketch^T\alpha \Big \},
\end{align}
and then outputting the function
\begin{align}
\label{EqnSketchedKRREstimate}
\fhat(\cdot) & \defn \frac{1}{\sqrt{\numobs}} \sum_{i=1}^\numobs
(\Sketch^T \alphahat)_i \KerFun{\cdot}{x_i}.
\end{align}
\end{subequations}
Note that the sketched program~\eqref{EqnSketchedKRR} is a quadratic
program in $\numproj$ dimensions: it takes as input the
$\numproj$-dimensional matrices $(\Sketch \KerMat^2 \Sketch^T, \,
\Sketch \KerMat \Sketch^T)$ and the $\numproj$-dimensional vector
$SKy$.  Consequently, it can be solved efficiently via QR
decomposition with computational complexity $\order(\numproj^3)$. Moreover, the computation of the sketched kernel matrix $\Sketch\KerMat=[\Sketch\KerMat_1,\ldots,\Sketch\KerMat_n]$ in the input can be parallellized across its columns.

In this paper, we analyze various forms of random sketch matrices
$\Sketch$.  Let us consider a few of them here.

\paragraph{Sub-Gaussian sketches:} We say
the row $s_i$ of the sketch matrix is zero-mean $1$-sub-Gaussian if
for any fixed unit vector $u \in \SPHERE{\numobs}$, we have
\begin{align*}
\mprob \big[|\inprod{u}{s_i}\geq t| \big]\leq 2e^{-\frac{\numobs
    \delta^2}{2}}\quad \mbox{for all $\delta \geq 0$.}
\end{align*}
Many standard choices of sketch matrices have i.i.d. $1$-sub-Gaussian
rows in this sense; examples include matrices with i.i.d. Gaussian
entries, i.i.d. Bernoulli entries, or random matrices with independent
rows drawn uniformly from a rescaled sphere.  For convenience, the
sub-Gaussian sketch matrices considered in this paper are all rescaled
so that their rows have the covariance matrix
$\frac{1}{\sqrt{\numproj}} I_{\numobs \times \numobs}$.

\paragraph{Randomized orthogonal system (ROS) sketches:} 
This class of sketches are based on randomly sampling and rescaling
the rows of a fixed orthonormal matrix $H \in \real^{\numobs \times
  \numobs}$.  Examples of such matrices include the discrete Fourier
transform (DFT) matrix, and the Hadamard matrix.  More specifically, a
ROS sketch matrix $\Sketch \in \real^{\numproj \times \numobs}$ is
formed with i.i.d.\!  rows of the form
\begin{align*}
s_i & = \sqrt{\frac{\numobs}{\numproj}} \RADE H^T p_i, \quad \mbox{for
  $i = 1, \ldots, \numproj$},
\end{align*}
where $\RADE$ is a random diagonal matrix whose entries are i.i.d.\!
Rademacher variables and $\{p_1,\ldots,p_m\}$ is a random subset of
$\numproj$ rows sampled uniformly from the $\numobs \times \numobs$
identity matrix without replacement. An advantage of using ROS
sketches is that for suitably chosen orthonormal matrices, including
the DFT and Hadamard cases among others, a matrix-vector product (say
of the form $\Sketch u$ for some vector $u \in \real^\numobs$) can be
computed in $\order(\numobs \log \numproj)$ time, as opposed to
$\order(\numobs \numproj)$ time required for the same operation with
generic dense sketches. For instance, see Ailon and
Liberty~\cite{Ailon08} and \cite{PilWai14a} for further details.  Throughout this paper, we
focus on ROS sketches based on orthonormal matrices $H$ with uniformly
bounded entries, meaning that $|H_{ij}| \leq \frac{c}{\sqrt{\numobs}}$
for all entries $(i,j)$.  This entrywise bound is satisfied by
Hadamard and DFT matrices, among others.

\paragraph{Sub-sampling sketches:}  This class of sketches 
are even simpler, based on sub-sampling the rows of the identity
matrix without replacement.  In particular, the sketch matrix $\Sketch
\in \real^{\numproj \times \numobs}$ has rows of the form $s_i =
\sqrt{\frac{\numobs}{\numproj}} \: p_i$, where the vectors $\{p_1,
\ldots, p_\numproj \}$ are drawn uniformly at random without
replacement from the $\numobs$-dimensional identity matrix.  It can be
understood as related to a ROS sketch, based on the identity matrix as
an orthonormal matrix, and not using the Rademacher randomization nor
satisfying the entrywise bound.  In Appendix~\ref{AppNystrom}, we show
that the sketched KRR estimate~\eqref{EqnSketchedKRR} based on a
sub-sampling sketch matrix is equivalent to the \NYS approximation.

%%%%%%%%%%%%%%%%%%%%%%%%%%%%%%%%%%%%%%%%%%%%%%%%%%%%%%%%%%%%%%%%%%%%

\subsection{Kernel complexity measures and statistical guarantees}
\label{SecKernelComplexity}

So as to set the stage for later results, let us characterize an
appropriate choice of the regularization parameter $\regpar$, and the
resulting bound on the prediction error $\|\fdagger -
\fstar\|_\numobs$.  Recall the empirical kernel matrix $\KerMat$
defined in the previous section: since it is symmetric and positive
definite, it has an eigendecomposition of the form \mbox{$\KerMat = U
  D U^T$,} where $U \in \real^{\numobs \times \numobs}$ is an
orthonormal matrix, and $D \in \real^{\numobs \times \numobs}$ is
diagonal with elements $\empkereig_1 \geq \empkereig_2 \geq \ldots
\geq \empkereig_\numobs \geq 0$.  Using these eigenvalues, consider
the \emph{kernel complexity function}
\begin{align}
\label{EqnDefnEmpKerComp}
\EmpKerComp(\delta) & = \sqrt{\frac{1}{\numobs} \sum_{j=1}^\numobs
  \min \{ \delta^2, \empkereig_j \}},
\end{align}
corresponding to a rescaled sum of the eigenvalues, truncated at level
$\delta^2$.  This function arises via analysis of the local Rademacher
complexity of the kernel class
(e.g.,~\cite{Bar05,Koltchinskii06,Mendelson02, RasWaiYu10b}).  For a
given kernel matrix and noise variance $\sigma > 0$, the
\emph{critical radius} is defined to be the smallest positive solution
$\delcrit > 0$ to the inequality
\begin{align}
\label{EqnDefnCritRad}
\frac{\EmpKerComp(\delta)}{\delta} & \leq \frac{\delta}{\sigma}.
\end{align}
Note that the existence and uniqueness of this critical radius is
guaranteed for any kernel class~\cite{Bar05}.

\paragraph{Bounds on ordinary KRR:}
The significance of the critical radius is that it can be used to
specify bounds on the prediction error in kernel ridge regression.
More precisely suppose that we compute the KRR estimate~\eqref{EqnKRR}
with any regularization parameter $\regpar \geq 2 \delcrit^2$.  Then
with probability at least $1 - c_1 e^{-c_2 \numobs \delcrit^2}$, we
are guaranteed that
\begin{align}
\label{EqnOriginalKRROptimality}
\|\fdagger - \fstar\|_\numobs^2 & \leq \cupper \; \big \{ \regparn +
\delcrit^2 \big \},
\end{align}
where $\cupper > 0$ is a universal constant (independent of $\numobs$,
$\sigma$ and the kernel).  This known result follows from standard
techniques in empirical process theory (e.g.,~\cite{vandeGeer,
  Bar05}); we also note that it can be obtained as a corollary of our
more general theorem on sketched KRR estimates to follow
(viz. Theorem~\ref{ThmMain0}). 

To illustrate, let us consider a few examples of reproducing kernel
Hilbert spaces, and compute the critical radius in different cases.
In working through these examples, so as to determine explicit rates,
we assume that the design points $\{x_i\}_{i=1}^\numobs$ are sampled
i.i.d. from some underlying distribution $\mprob$, and we make use of
the useful fact that, up to constant factors, we can always work with
the population-level kernel complexity function
\begin{align}
\label{EqnDefnPopKerComp}
\PopKerComp(\delta) & = \sqrt{\frac{1}{\numobs} \sum_{j=1}^\infty
  \min \{ \delta^2, \popkereig_j \}},
\end{align}
where $\{\popkereig_j\}_{j=1}^\infty$ are the eigenvalues of the
kernel integral operator (assumed to be uniformly bounded).  This
equivalence follows from standard results on the population and
empirical Rademacher complexities~\cite{Mendelson02,Bar05}.

\bexs[Polynomial kernel]
\label{ExaPolynomialIntro}
For some integer $\polydeg \geq 1$, consider the kernel function on
$[0,1] \times [0,1]$ given by $\KerFunPoly{u}{v} = \big(1 +
\inprod{u}{v} \big)^\polydeg$.  For $\polydeg = 1$, it generates the
class of all linear functions of the form $f(x) = a_0 + a_1 x$ for
some scalars $(a_0, a_1)$, and corresponds to a linear kernel.  More
generally, for larger integers $\polydeg$, it generates the class of
all polynomial functions of degree at most $\polydeg$---that is,
functions of the form $f(x) = \sum_{j=0}^{\polydeg} a_j x^{j}$.

Let us now compute a bound on the critical radius $\delcrit$. It is
straightforward to show that the polynomial kernel is of finite rank
at most $\polydeg + 1$, meaning that the kernel matrix $\KerMat$
always has at most $\min \{\polydeg + 1, \numobs \}$ non-zero
eigenvalues.  Consequently, as long $\numobs > \polydeg + 1$, there is
a universal constant $c$ such that
\begin{align*}
\EmpKerComp(\delta) & \leq c \sqrt{ \frac{\polydeg + 1 }{\numobs}
  \delta},
\end{align*}
which implies that $\delcrit^2 \precsim \sigma^2 \frac{\polydeg + 1
}{\numobs}$.  Consequently, we conclude that the KRR estimate
satisifes the bound $\|\fhat - \fstar\|_\numobs^2 \precsim \sigma^2
\frac{\polydeg + 1}{\numobs}$ with high probability.  Note that this
bound is intuitive, since a polynomial of degree $\polydeg$ has
$\polydeg + 1$ free parameters.

\eexs

\bexs[Gaussian kernel]
\label{ExaGaussianIntro}
The Gaussian kernel with bandwidth $h > 0$ takes the form
\mbox{$\KerFunGauss{u}{v} = e^{-\frac{1}{2h^2}(u-v)^2}$.}  When
defined with respect to Lebesgue measure on the real line, the
eigenvalues of the kernel integral operator scale as $\mu_j \asymp
\exp(-\pi h^2 j^2)$ as $j \to \infty$. Based on this fact, it can be
shown that the critical radius scales as $\delcrit^2 \asymp
\frac{\sigma^2}{\numobs} \sqrt{\log
  \big(\frac{\numobs}{\sigma^2}\big)}$.  Thus, even though the
Gaussian kernel is non-parametric (since it cannot be specified by a
fixed number of parametrers), it is still a relatively small function
class.
\eexs

\bexs[First-order Sobolev space]
\label{ExaLipschitzIntro}
As a final example, consider the kernel defined on the unit square
$[0,1] \times [0,1]$ given by $\KerFunLip{u}{v} = \min \{u, v\}$.  It
generates the function class
\begin{equation}
\label{EqnFirstOrderSob}
\begin{aligned}
\Hil^{1}[0,1]  = \Big\{&\,f:[0,1] \to \real\, \mid \, \mbox{$f(0) = 0$},\\
  &\mbox{and $f$ is abs. cts. with $\int_0^1[f'(x)]^2\, dx < \infty$} \big\},
\end{aligned}
\end{equation}
a class that contains all Lipschitz functions on the unit interval
$[0,1]$.  Roughly speaking, we can think of the first-order Sobolev
class as functions that are almost everywhere differentiable with
derivative in $L^2[0,1]$.  Note that this is a much larger kernel
class than the Gaussian kernel class.  The first-order Sobolev space
can be generalized to higher order Sobolev spaces, in which functions
have additional smoothness.  See the book~\cite{Gu02} for further
details on these and other reproducing kernel Hilbert spaces.

If the kernel integral operator is defined with respect to Lebesgue
measure on the unit interval, then the population level eigenvalues
are given by $\mu_j = \big( \frac{2}{(2j-1)^2 \pi} \big)^2$ for $j =
1, 2, \ldots$.  Given this relation, some calculation shows that the
critical radius scales as $\delcrit^2 \asymp
\big(\frac{\sigma^2}{\numobs} \big)^{2/3}$.  This is the familiar
minimax risk for estimating Lipschitz functions in one
dimension~\cite{Stone82}.
\eexs
%

%

%%%%%%%%%%%%%%%%%%%%%%%%%%%%%%%%%%%%%%%%%%%%%%%%%%%%%%%%%%%%%%%%%%%%%

\paragraph{Lower bounds for non-parametric regression:}

For future reference, it is also convenient to provide a lower bound
on the prediction error achievable by \emph{any estimator}.  In order
to do so, we first define the \emph{statistical dimension} of the
kernel as
\begin{align}
\label{EqnDefnStatdim}
\effdim & \defn \arg \min_{j = 1, \ldots, \numobs} \big \{
\empkereig_j \leq \delcrit^2 \},
\end{align}
and $\effdim = \numobs$ if no such index $j$ exists.  By definition,
we are guaranteed that $\empkereig_{j} > \delcrit^2$ for all $j \in
\{1, 2, \ldots, \effdim \}$. In terms of this statistical dimension,
we have
\begin{align*}
\EmpKerComp(\delcrit) & = \Big[ \frac{\effdim}{\numobs} \delcrit^2 +
  \frac{1}{\numobs} \sum_{j = \effdim+1}^\numobs \empkereig_j
  \Big]^{1/2},
\end{align*}
showing that the statistical dimension controls a type of bias-variance
tradeoff.   

It is reasonable to expect that the critical rate $\delcrit$ should be
related to the statistical dimension as $\delcrit^2 \asymp
\frac{\sigma^2\effdim}{\numobs}$.  This scaling relation holds whenever the
tail sum satisfies a bound of the form $\sum_{j = \effdim + 1}^\numobs
\empkereig_j \precsim \effdim \delcrit^2$.  Although it is possible to
construct pathological examples in which this scaling relation does
not hold, it is true for most kernels of interest, including all
examples considered in this paper.  For any such regular kernel, the
critical radius provides a fundamental lower bound on the performance
of \emph{any estimator}, as summarized in the following theorem:

\btheos[Critical radius and minimax risk]
\label{ThmLowerBound}
Given $\numobs$ i.i.d. samples $\{(y_i, x_i)\}_{i=1}^\numobs$ from the
standard non-parametric regression model over any regular kernel
class, any estimator $\ftil$ has prediction error lower bounded as
\begin{align}
\label{EqnLowerBound}
\sup_{\|\fstar\|_\Hil \leq 1} \Exs \| \flest - \fstar\|_\numobs^2 &
\geq \clower \delcrit^2,
\end{align}
where $\clower > 0$ is a numerical constant, and $\delcrit$ is the
critical radius~\eqref{EqnDefnCritRad}.
\etheos
\noindent The proof of this claim, provided in
Appendix~\ref{SecLowerBound}, is based on a standard applicaton of
Fano's inequality, combined with a random packing argument.  It
establishes that the critical radius is a fundamental quantity,
corresponding to the appropriate benchmark to which sketched kernel
regression estimates should be compared.

%%%%%%%%%%%%%%%%%%%%%%%%%%%%%%%%%%%%%%%%%%%%%%%%%%%%%%%%%%%%%%%%%%%%%%%%%%%%%

\section{Main results and their consequences}
\label{SecMain}

We now turn to statements of our main theorems on kernel sketching, as
well as a discussion of some of their consequences.  We first
introduce the notion of a $\KerMat$-satisfiable sketch matrix, and
then show (in Theorem~\ref{ThmMain0}) that any sketched KRR estimate
based on a $\KerMat$-satisfiable sketch also achieves the minimax
risk.  We illustrate this achievable result with several corollaries
for different types of randomized sketches.  For Gaussian and ROS
sketches, we show that choosing the sketch dimension proportional to
the statistical dimension of the kernel (with additional log factors
in the ROS case) is sufficient to guarantee that the resulting sketch
will be $\KerMat$-satisfiable with high probability. In addition, we
illustrate the sharpness of our theoretical predictions via some
experimental simulations.

%%%%%%%%%%%%%%%%%%%%%%%%%%%%%%%%%%%%%%%%%%%%%%%%%%%%%%%%%%%%%%%%%%%%%%%%%%

\subsection{General conditions for sketched kernel optimality}

Recall the definition~\eqref{EqnDefnStatdim} of the statistical
dimension $\statdim$, and consider the eigendecomposition
\mbox{$\KerMat = U D U^T$} of the kernel matrix, where $U \in
\real^{\numobs \times \numobs}$ is an orthonormal matrix of
eigenvectors, and $D = \diag \{\empkereig_1, \ldots, \empkereig_\numobs \}$ is
a diagonal matrix of eigenvalues.  Let $U_1 \in \real^{\numobs
  \times\effdim}$ denote the left block of $U$, and similarly, $U_2
\in \real^{\numobs \times (\numobs - \effdim)}$ denote the right
block.  Note that the columns of the left block $U_1$ correspond to
the eigenvectors of $K$ associated with the leading $\statdim$
eigenvalues, whereas the columns of the right block $U_2$ correspond
to the eigenvectors associated with the remaining $\numobs - \statdim$
smallest eigenvalues.  Intuitively, a sketch matrix \mbox{$\Sketch \in
  \real^{\numproj \times \numobs}$} is ``good'' if the sub-matrix
$\Sketch U_1 \in \real^{\numproj \times \statdim}$ is relatively close
to an isometry, whereas the sub-matrix $\Sketch U_2 \in
\real^{\numproj \times (\numobs - \statdim)}$ has a relatively small
operator norm.

This intuition can be formalized in the following way. For a given
kernel matrix $\KerMat$, a sketch matrix $\Sketch$ is said to be
\emph{$\KerMat$-satisfiable} if there is a universal constant
$\PRECON$ such that
\begin{align}
\label{EqnKerSat}
\opnorm{(\Sketch U_1)^T \Sketch U_1 - I_{\effdim}}\leq 1/2, \quad
\mbox{and} \quad \opnorm{\Sketch U_2 \, D_2^{1/2}} \leq \PRECON \;
\delcrit,
\end{align}
where $D_2 = \diag \{ \empkereig_{\statdim+1}, \ldots,  \empkereig_{\numobs} \}$.

Given this definition, the following theorem shows that any sketched
KRR estimate based on a $\KerMat$-satisfiable matrix achieves the
minimax risk (with high probability over the noise in the observation
model):
\btheos[Upper bound]
\label{ThmMain0}
Given $\numobs$ i.i.d. samples $\{(y_i, x_i)\}_{i=1}^\numobs$ from the
standard non-parametric regression model, consider the sketched KRR
problem~\eqref{EqnSketchedKRR} based on a $\KerMat$-satisfiable sketch
matrix $\Sketch$.  Then any for $\regparn \geq 2\delcrit^2$, the
sketched regression estimate $\fhat$ from
equation~\eqref{EqnSketchedKRREstimate} satisfies the bound
\begin{align*}
\|\fhat - \fstar\|_\numobs^2 & \leq \cupper \, \big \{ \regparn +
\delcrit^2 \big \}
\end{align*}
with probability greater than $1 - c_1 \CEXP{-c_2\numobs \delcrit^2}$.
\etheos

We emphasize that in the case of fixed design regression and for a
fixed sketch matrix, the $\KerMat$-satisfiable condition on the sketch
matrix $\Sketch$ is a deterministic statement: apart from the sketch
matrix, it only depends on the properties of the kernel function
$\mathcal{K}$ and design variables $\{x_i\}_{i=1}^\numobs$.  Thus,
when using randomized sketches, the algorithmic randomness can be
completely decoupled from the randomness in the noisy observation
model~\eqref{EqnModel}.

\paragraph{Proof intuition:}
The proof of Theorem~\ref{ThmMain0} is given in
Section~\ref{SecProofThmMain0}.  At a high-level, it is based on an
upper bound on the prediction error $\|\fhat - \fstar\|_\numobs^2$
that involves two sources of error: the \emph{approximation error}
associated with solving a zero-noise version of the KRR problem in the
projected $\numproj$-dimensional space, and the \emph{estimation
  error} between the noiseless and noisy versions of the projected
problem.  In more detail, letting $\FstarEval \defn \{\fstar(x_1),
\ldots, \fstar(x_\numobs) \}$ denote the vector of function
evaluations defined by $\fstar$, consider the quadratic program
\begin{align}
\label{EqnDefnAlphaDagger}
\alphadagger & \defn \arg \min_{\alpha \in \real^\numproj} \Big \{
\frac{1}{2 \numobs} \|\FstarEval - \sqrt{\numobs}K \Sketch^T
\alpha\|_2^2 + \regparn \|\sqrt{K} \Sketch^T \alpha\|_2^2 \Big\},
\end{align}
as well as the associated fitted function $\fdagger =
\frac{1}{\sqrt{\numobs}} \sum_{i=1}^\numobs \alphadagger_i
\KerFun{\cdot}{x_i}$.  The vector $\alphadagger \in \real^\numproj$ is
the solution of the sketched problem in the case of zero noise,
whereas the fitted function $\fdagger$ corresponds to the best
penalized approximation of $\fstar$ within the range space of
$\Sketch^T$.

Given this definition, we then have the elementary inequality
\begin{align}
\label{EqnDecomposition}
\frac{1}{2} \|\fhat - \fstar\|_\numobs^2 & \leq \underbrace{\|\fdagger
  - \fstar\|_\numobs^2}_{\mbox{Approximation error}} +
\underbrace{\|\fdagger - \fhat\|_\numobs^2}_{\mbox{Estimation error}}.
\end{align}
For a fixed sketch matrix, the approximation error term is
deterministic: it corresponds to the error induced by approximating
$\fstar$ over the range space of $\Sketch^T$.  On the other hand, the
estimation error depends both on the sketch matrix and the observation
noise.  In Section~\ref{SecProofThmMain0}, we state and prove two
lemmas that control the approximation and error terms respectively.

As a corollary, Theorem~\ref{ThmMain0} implies the stated upper
bound~\eqref{EqnOriginalKRROptimality} on the prediction error of the
original (unsketched) KRR estimate~\eqref{EqnKRR}.  Indeed, this
estimator can be obtained using the ``sketch matrix'' $\Sketch =
I_{\numobs \times \numobs}$, which is easily seen to be
$\KerMat$-satisfiable.  In practice, however, we are interested in
$\numproj \times \numobs$ sketch matrices with $\numproj \ll \numobs$,
so as to achieve computational savings.  In particular, a natural
conjecture is that it should be possible to efficiently generate
$\KerMat$-satisfiable sketch matrices with the projection dimension
$\numproj$ proportional to the statistical dimension $\statdim$ of the
kernel.  Of course, one such $\KerMat$-satisfiable matrix is given by
$\Sketch = U_1^T \in \real^{\statdim \times \numobs}$, but it is not
easy to generate, since it requires computing the eigendecomposition
of $\KerMat$.  Nonetheless, as we now show, there are various
randomized constructions that lead to $\KerMat$-satisfiable sketch
matrices with high probability.

%%%%%%%%%%%%%%%%%%%%%%%%%%%%%%%%%%%%%%%%%%%%%%%%%%%%%%%%%%%%%%%%%%%%%%%%%
\subsection{Corollaries for randomized sketches}

When combined with additional probabilistic analysis,
Theorem~\ref{ThmMain0} implies that various forms of randomized
sketches achieve the minimax risk using a sketch dimension
proportional to the statistical dimension $\statdim$.  Here we analyze
the Gaussian and ROS families of random sketches, as previously
defined in Section~\ref{SecKernelSketchBackground}.  Throughout our
analysis, we require that the sketch dimension satisfies a lower obund
of the form
\begin{subequations}
\begin{align}
\label{EqnDefnKgood}
\numproj & \geq \begin{cases} \PRECON \: \statdim & \mbox{for Gaussian
    sketches, and}\\ \PRECON \: \statdim \log^4(\numobs) & \mbox{for
    ROS sketches,}
\end{cases}
\end{align}
where $\statdim$ is the \emph{statistical dimension} as previously
defined in equation~\eqref{EqnDefnStatdim}.  Here it should be
understood that the constant $\PRECON$ can be chosen sufficiently
large (but finite).  In addition, for the purposes of stating high
probability results, we define the function
\begin{align}
\label{EqnDefnPhiProb}
\PHIPROB & \defn
\begin{cases} c_1 \CEXP{-c_2\numproj } &
  \mbox{for Gaussian sketches, and} \\
c_1 \biggr[ \CEXP{- c_2 \frac{\numproj}{\statdim \log^2(\numobs)}} +
  \CEXP{-c_2 \statdim \log^2(\numobs)} \biggr] & \mbox{for ROS
  sketches},
\end{cases}
\end{align}
\end{subequations}
where $c_1, c_2$ are universal constants.  With this notation, the
following result provides a high probability guarantee for both
Gaussian and ROS sketches:
\bcors[Guarantees for Gaussian and ROS sketches]
\label{CorSubgaussian}
Given $\numobs$ i.i.d. samples $\{(y_i, x_i)\}_{i=1}^\numobs$ from the
standard non-parametric regression model~\eqref{EqnModel}, consider
the sketched KRR problem~\eqref{EqnSketchedKRR} based on a sketch
dimension $\numproj$ satisfying the lower
bound~\eqref{EqnDefnKgood}. Then there is a universal constant
$\cupper'$ such that for any $\regparn \geq 2\delcrit^2$, the sketched
regression estimate~\eqref{EqnSketchedKRREstimate} satisfies the bound
\begin{align*}
\|\fhat - \fstar\|_\numobs^2 & \leq \cupper' \, \big \{ \regparn +
\delcrit^2 \big \}
\end{align*}
with probability greater than $1 - \PHIPROB - c_3 \CEXP{-c_4 \numobs
  \delcrit^2}$.
\ecors

\noindent In order to illustrate Corollary~\ref{CorSubgaussian}, let
us return to the three examples previously discussed in
Section~\ref{SecKernelComplexity}. To be concrete, we derive the
consequences for Gaussian sketches, noting that ROS sketches incur
only an additional $\log^4(\numobs)$ overhead.

\begin{itemize}
\item for the $\polydeg^{th}$-order polynomial kernel from
  Example~\ref{ExaPolynomialIntro}, the statistical dimension
  $\statdim$ for any sample size $\numobs$ is at most $\polydeg + 1$,
  so that a sketch size of order $\polydeg + 1$ is sufficient.  This
  is a very special case, since the kernel is finite rank and so the
  required sketch dimension has no dependence on the sample size.
\item for the Gaussian kernel from Example~\ref{ExaGaussianIntro}, the
  statistical dimension satisfies the scaling $\statdim \asymp
  \sqrt{\log \numobs}$, so that it suffices to take a sketch dimension
  scaling logarithmically with the sample size.
\item for the first-order Sobolev kernel from
  Example~\ref{ExaLipschitzIntro} , the statistical dimension scales
  as $\statdim \asymp \numobs^{1/3}$, so that a sketch dimension
  scaling as the cube root of the sample size is required.
\end{itemize}

\begin{figure}[h]
\begin{center}
\begin{tabular}{ccc}
\widgraph{.45\textwidth}{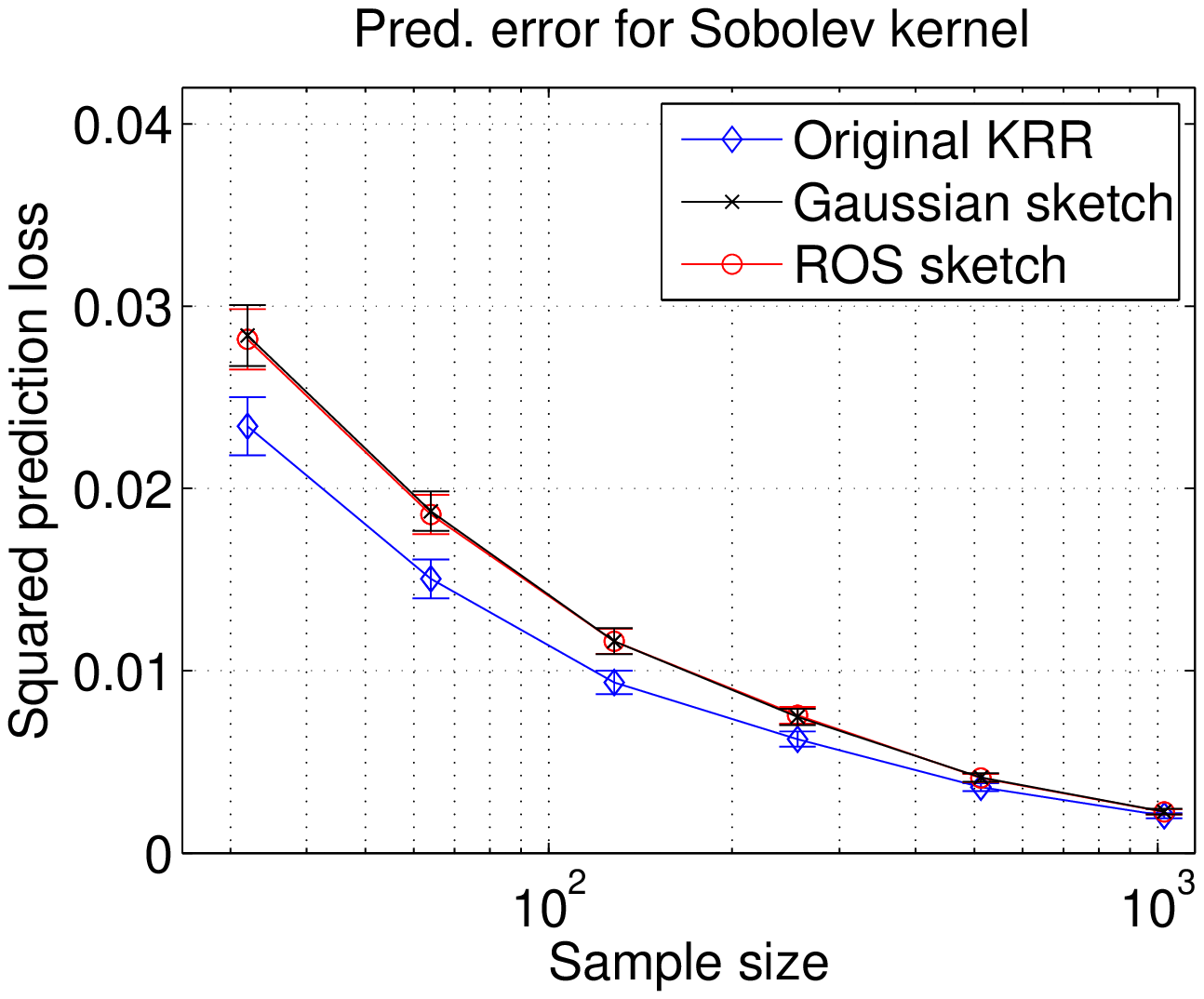} & &
\widgraph{.45\textwidth}{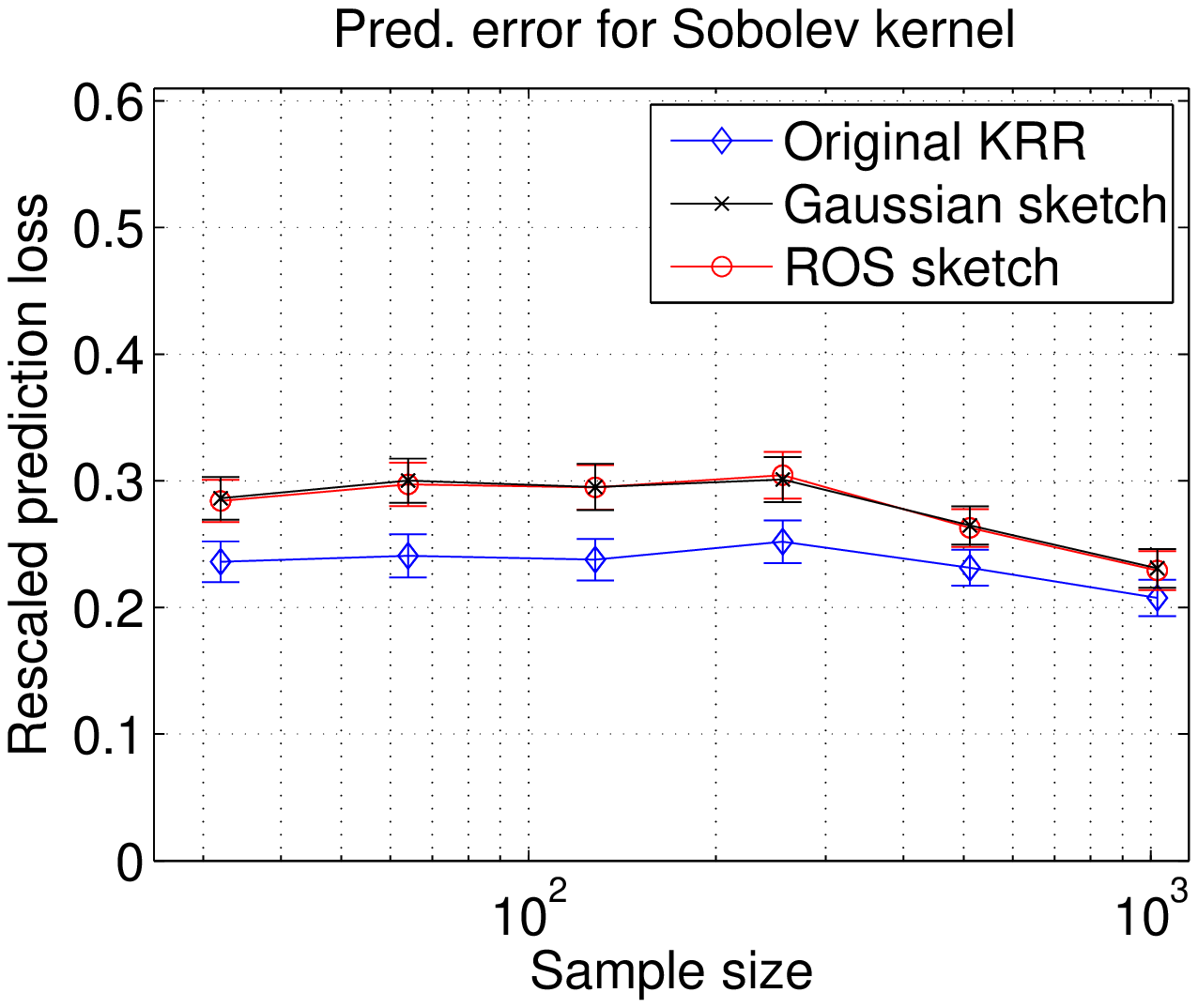} \\
(a) & & (b) 
\end{tabular}
\end{center}
\caption{Prediction error versus sample size for original KRR,
  Gaussian sketch, and ROS sketches for the Sobolev one kernel for the
  function $\fstar(x) = |x + 0.5| - 0.5$. In all cases, each point
  corresponds to the average of $100$ trials, with standard errors
  also shown.  (a) Squared prediction error
  $\|\fhat-\fstar\|_\numobs^2$ versus the sample size $\numobs \in
  \{32,64,128,256,1024\}$ for projection dimension $\numproj=\lceil
  \numobs^{1/3} \rceil$.  (b) Rescaled prediction error $\numobs^{2/3}
  \|\fhat - \fstar\|_\numobs^2$ versus the sample size. }
\label{FigSimpleSob}
\end{figure}

In order to illustrate these theoretical predictions, we performed
some simulations.  Beginning with the Sobolev kernel $\KerFunLip{u}{v}
= \min \{u,v\}$ on the unit square, as introduced in
Example~\ref{ExaLipschitzIntro}, we generated $\numobs$ i.i.d. samples
from the model ~\eqref{EqnModel} with noise standard deviation $\sigma
= 1$, the unknown regression function
\begin{align}
\fstar(x) & = |x + 0.5| - 0.5,
\end{align} 
and uniformly spaced design points $x_i=\frac{i}{n}$ for $i=1,\ldots,
\numobs$.  By construction, the function $\fstar$ belongs to the
first-order Sobolev space with $\|\fstar\|_\Hil = 1$.  As suggested by
our theory for the Sobolev kernel, we set the projection dimension
$\numproj = \lceil n^{1/3} \rceil$, and then solved the sketched
version of kernel ridge regression, for both Gaussian sketches and ROS
sketches based on the fast Hadamard transform.  We performed
simulations for $\numobs$ in the set $\{32, 64, 128, 256, 512, 1024
\}$ so as to study scaling with the sample size.  As noted above, our
theory predicts that the squared prediction loss
$\|\fhat-\fstar\|_\numobs^2$ should tend to zero at the same rate
$\numobs^{-2/3}$ as that of the unsketched estimator $\fdagger$.
Figure~\ref{FigSimpleSob} confirms this theoretical prediction.  In
panel (a), we plot the squared prediction error versus the sample
size, showing that all three curves (original, Gaussian sketch and ROS
sketch) tend to zero.  Panel (b) plots the \emph{rescaled} prediction
error $\numobs^{2/3} \|\fhat - \fstar\|_\numobs^2$ versus the sample
size, with the relative flatness of these curves confirming the
$n^{-2/3}$ decay predicted by our theory.

\begin{figure}[h]
\begin{center}
\begin{tabular}{ccc}
\widgraph{.45\textwidth}{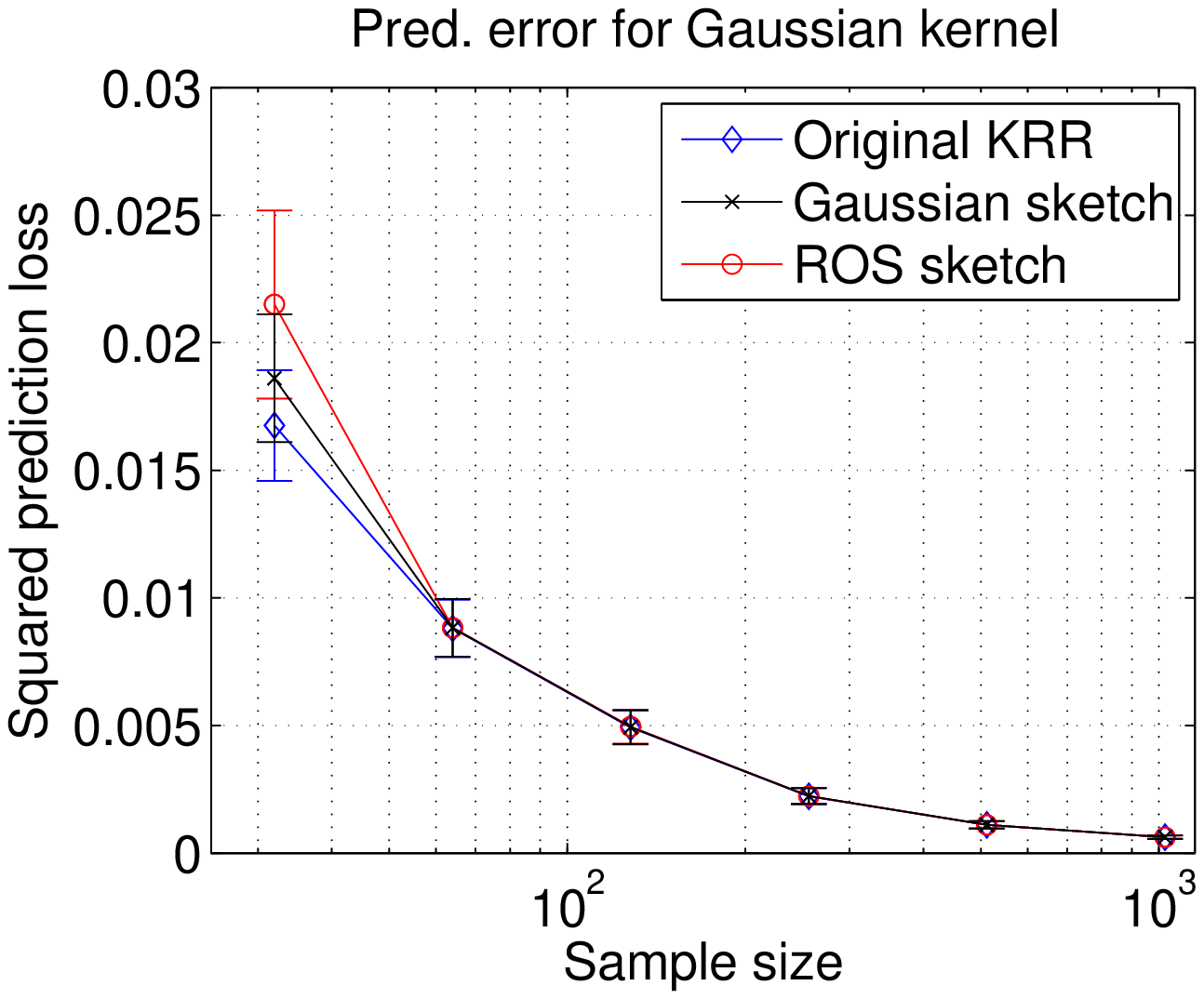} & & 
\widgraph{.45\textwidth}{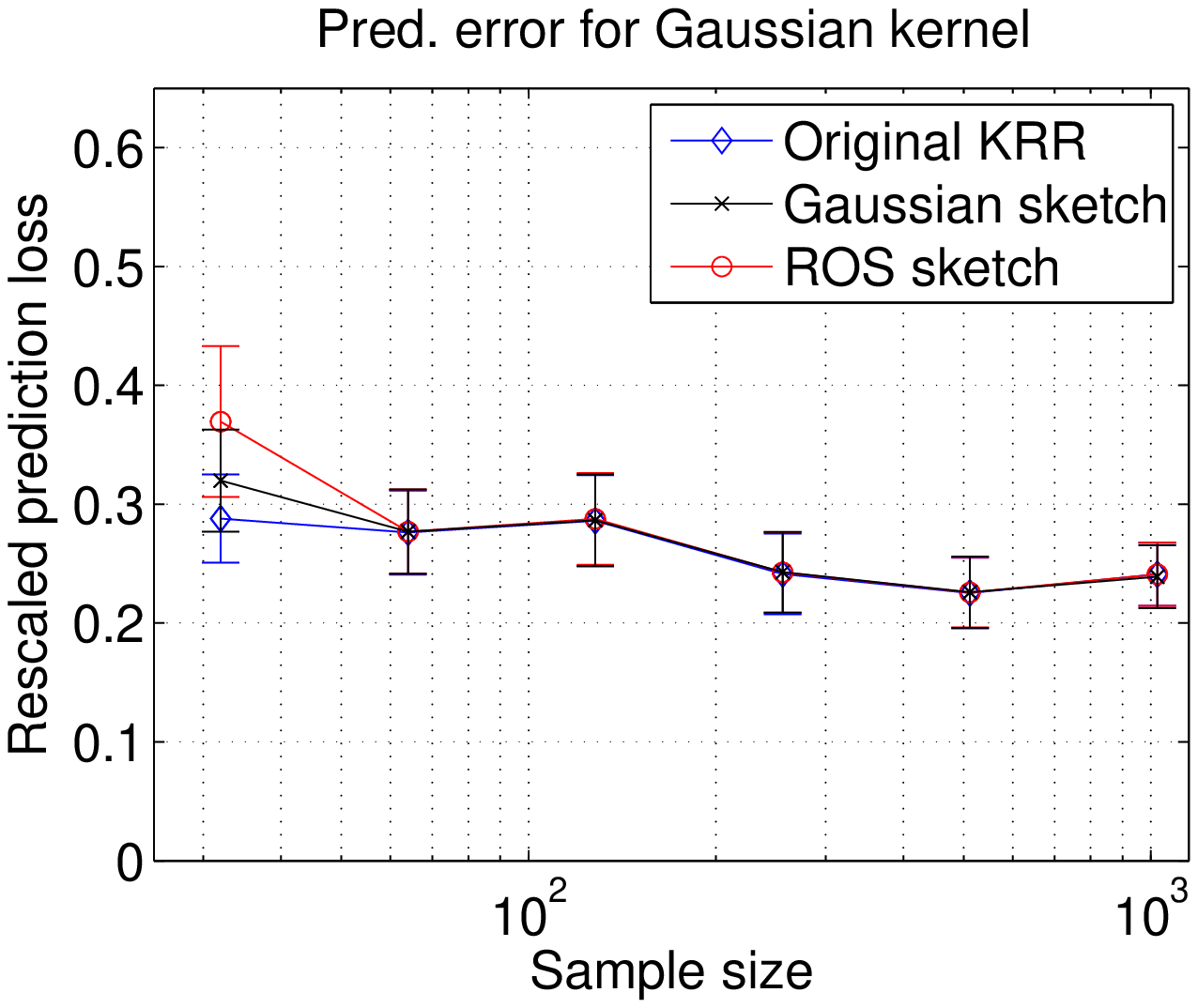} \\
(a) & & (b)
\end{tabular}
\end{center}
\caption{Prediction error versus sample size for original KRR,
  Gaussian sketch, and ROS sketches for the Gaussian kernel with the
  function $\fstar(x) = -1 + 2 x^2$. In all cases, each point
  corresponds to the average of $100$ trials, with standard errors
  also shown.  (a) Squared prediction error
  $\|\fhat-\fstar\|_\numobs^2$ versus the sample size $\numobs \in
  \{32,64,128,256,1024\}$ for projection dimension $\numproj= \lceil
  1.25 \sqrt{\log \numobs} \rceil$.  (b) Rescaled prediction error
  $\frac{\numobs}{\sqrt{\log \numobs}} \|\fhat - \fstar\|_\numobs^2$
  versus the sample size.  }
\label{FigSimpleGauss}
\end{figure}

In our second experiment, we repeated the same set of simulations this
time for the Gaussian kernel $\KerFunGauss{u}{v} = e^{-\frac{1}{2 h^2}
  (u -v)^2}$ with bandwidth $h = 0.25$, and the function
\mbox{$\fstar(x) = -1 + 2x^2$.}  In this case, as suggested by our
theory, we choose the sketch dimension $\numproj = \lceil 1.25
\sqrt{\log \numobs} \rceil$.  Figure~\ref{FigSimpleGauss} shows the
  same types of plots with the prediction error.  In this case, we
  expect that the squared prediction error will decay at the rate
  $\frac{\sqrt{\log \numobs}}{\numobs}$.  This prediction is confirmed
  by the plot in panel (b), showing that the rescaled error
  $\frac{\numobs}{\sqrt{\log \numobs}} \|\fhat - \fstar\|_\numobs^2$,
  when plotted versus the sample size, remains relatively constant
  over a wide range.

%%%%%%%%%%%%%%%%%%%%%%%%%%%%%%%%%%%%%%%%%%%%%%%%%%%%%%%%%%%%%%%%%%%%%%%%%

%%%%%%%%%%%%%%%%%%%%%%%%%%%%%%%%%%%%%%%%%%%%%%%%%%%%%%%%%%%%%%%%%

\subsection{Comparison with Nystr\"{o}m-based approaches}
\label{SecNystrom}

It is interesting to compare the convergence rate and computational
complexity of our methods with guarantees based on the \NYS
approximation.  As shown in Appendix~\ref{AppNystrom}, this \NYS
approximation approach can be understood as a particular form of our
sketched estimate, one in which the sketch corresponds to a random
row-sampling matrix.

Bach~\cite{Bach12} analyzed the prediction error of the \NYS
approximation to KRR based on uniformly sampling a subset of
$\pcol$-columns of the kernel matrix $\KerMat$, leading to an overall
computational complexity of $\order(\numobs p^2)$.  In order for the
approximation to match the performance of KRR, the number of sampled
columns must be lower bounded as
\begin{align*}
\pcol & \succsim \numobs \|\text{diag}(\KerMat (\KerMat +
\lambda_\numobs I)^{-1})\|_{\infty} \log \numobs,
\end{align*}
a quantity which can be substantially larger than the statistical
dimension required by our methods.  Moreover, as shown in the
following example, there are many classes of kernel matrices for which
the performance of the \NYS approximation will be poor.

\bexs[Failure of \NYS approximation]
\label{ExaFailure}
Given a sketch dimension $\numproj \leq n \log 2$, consider an
empirical kernel matrix $\KerMat$ that has a block diagonal form
$\mbox{diag}(K_1,K_2)$, where $K_1\in\real^{(\numobs-k)\times
  (\numobs-k)}$ and $K_2\in\real^{k\times k}$ for any integer $k \leq
\frac{\numobs}{\numproj} \log 2$. Then the probability of not sampling
any of the last $k$ columns/rows is at least
$1-(1-k/\numobs)^\numproj\geq 1-e^{-k\numproj/\numobs}\geq 1/2$. This
means that with probability at least $1/2$, the sub-sampling sketch
matrix can be expressed as $\Sketch =(\Sketch_1,0)$, where
$\Sketch_1\in\real^{\numproj\times (\numobs-k)}$. Under such an event,
the sketched KRR~\eqref{EqnSketchedKRR} takes on a degenerate form,
namely
\begin{align*}
\alphahat & = \arg \min_{\theta \in \real^\numproj} \Big \{
\frac{1}{2} \alpha^T \Sketch_1 K_1^2\Sketch_1^T \alpha -
\alpha^T\Sketch_1\, \frac{K_1 y_1}{\sqrt{\numobs}} + \lambda_\numobs
\alpha^T \Sketch_1 K_1\Sketch_1^T\alpha \Big \},
\end{align*}
and objective that depends only on the first $\numobs-k$ observations.
Since the values of the last $k$ observations can be arbitrary, this
degeneracy has the potential to lead to substantial approximation error.
\eexs
%

%\vsmall

\begin{figure}[h!]
\begin{center}
\begin{tabular}{ccc}
\widgraph{.45\textwidth}{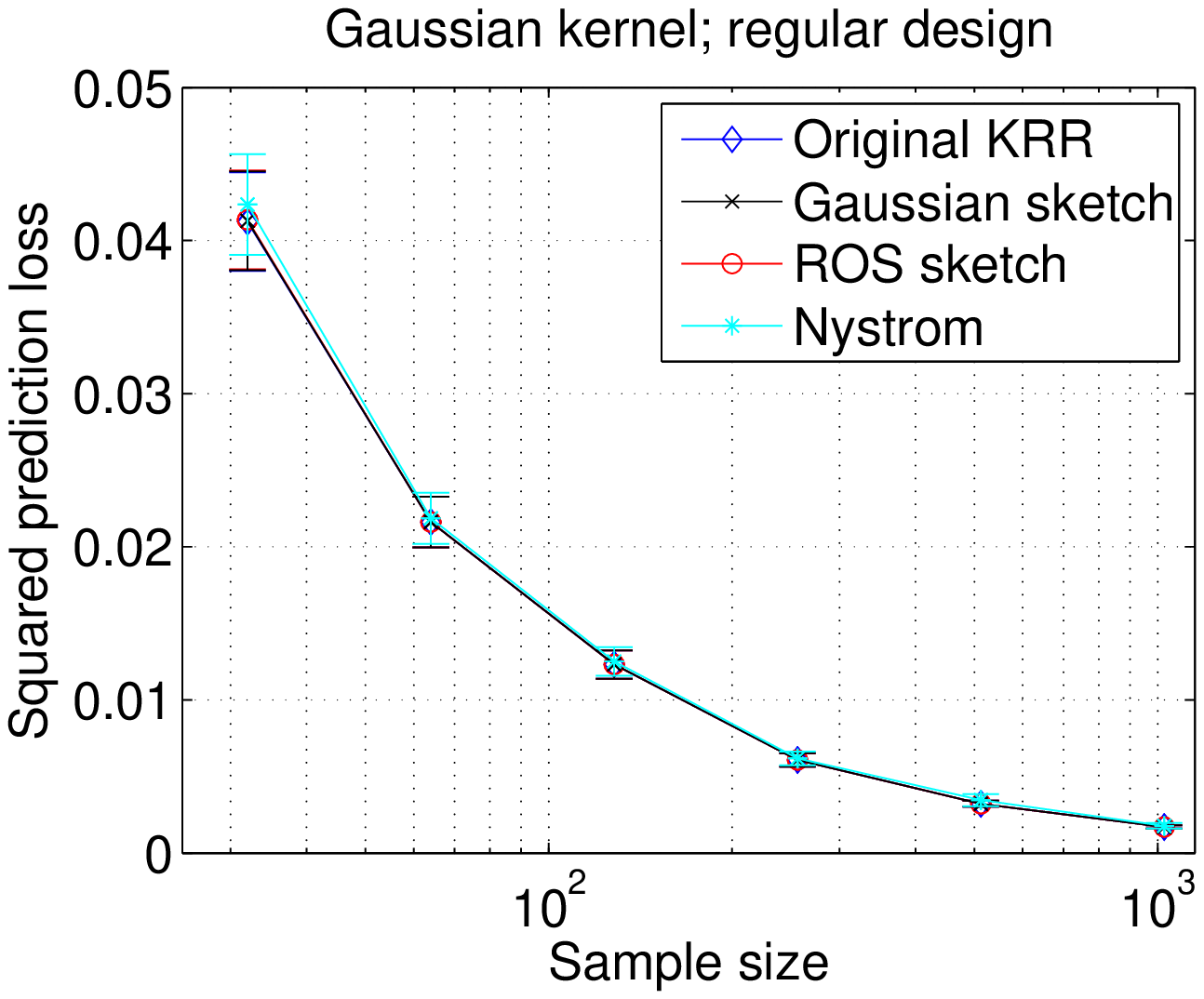} & & 
\widgraph{.45\textwidth}{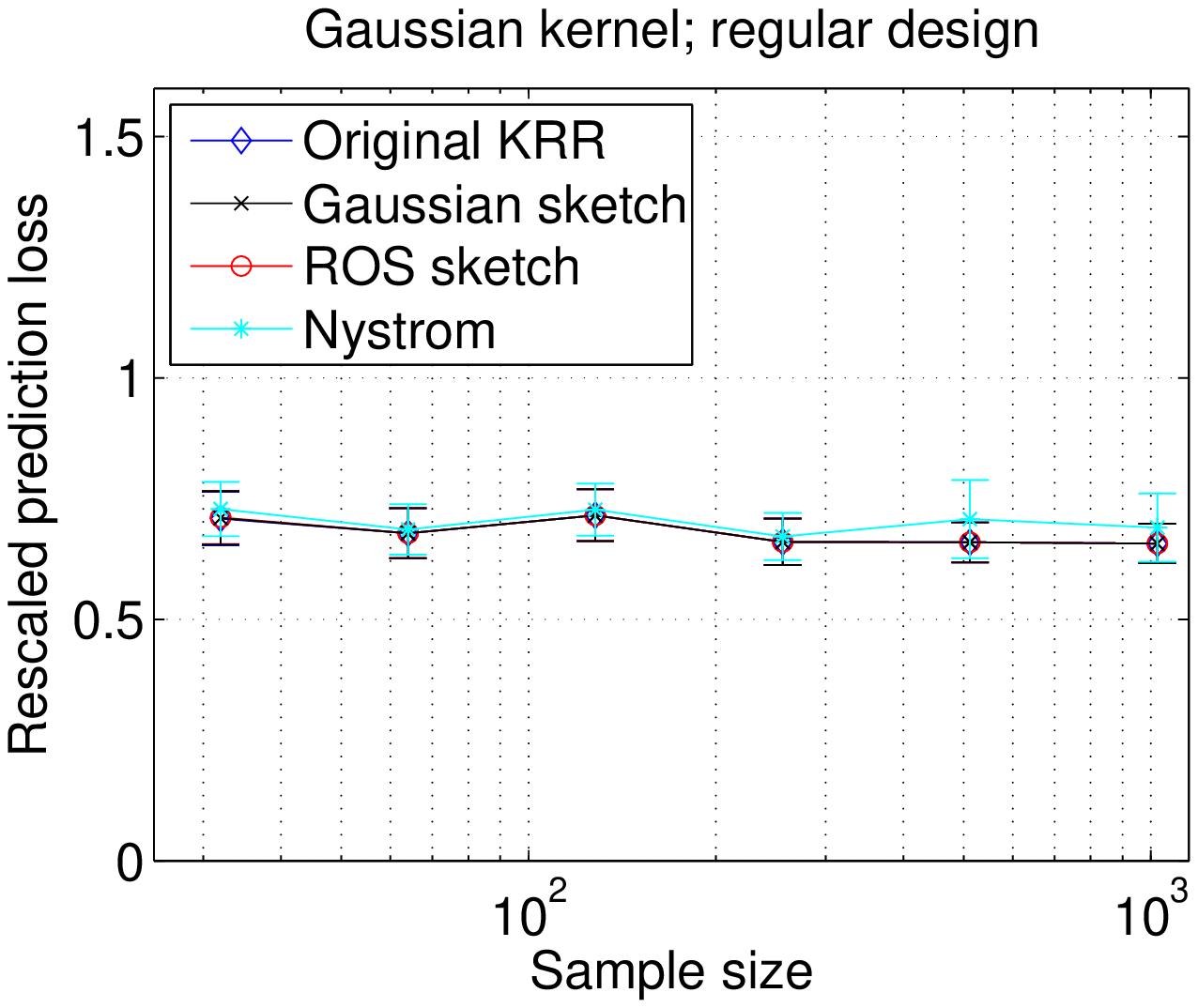} \\
(a) & & (b) \\
\widgraph{.45\textwidth}{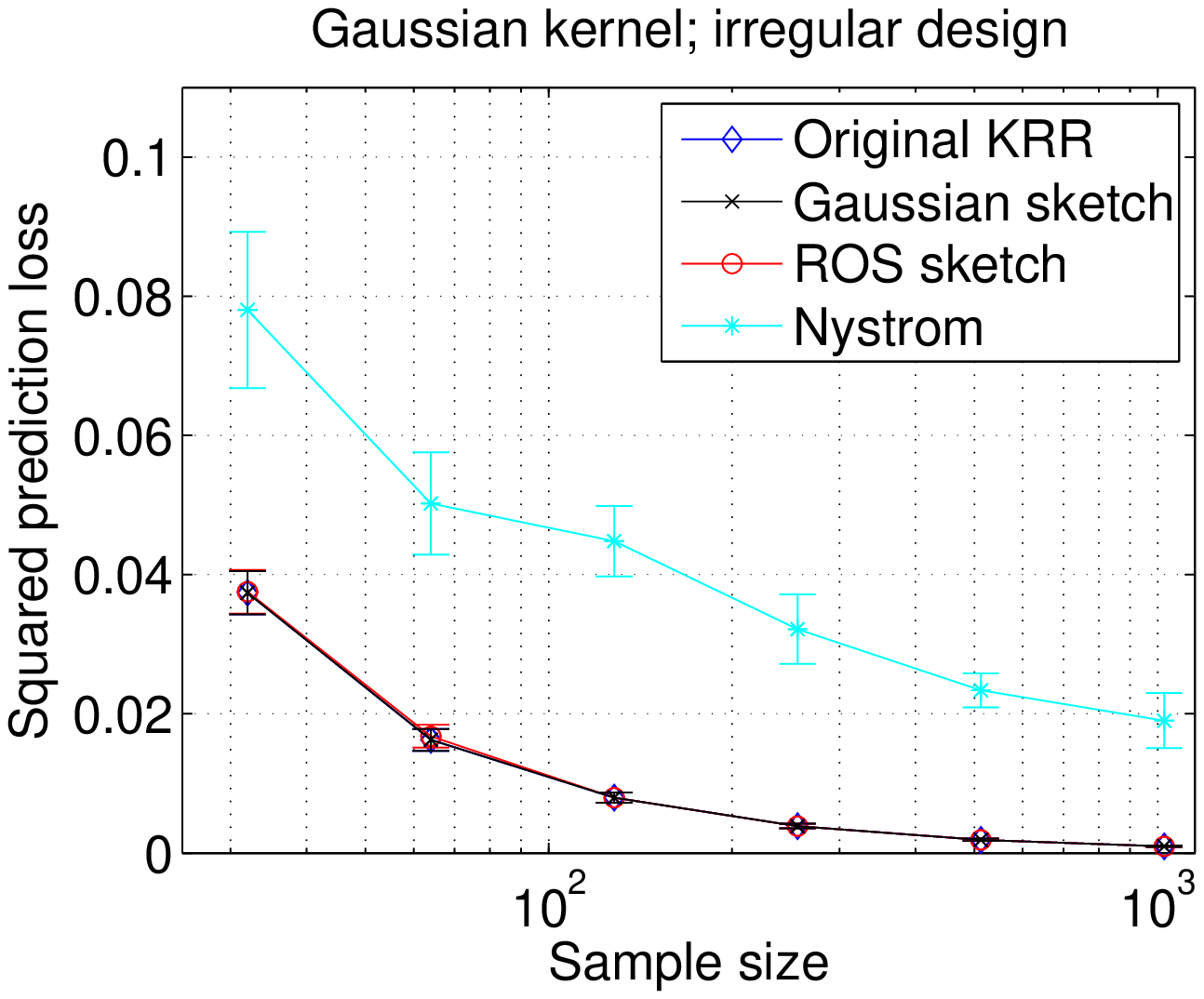} & & 
\widgraph{.45\textwidth}{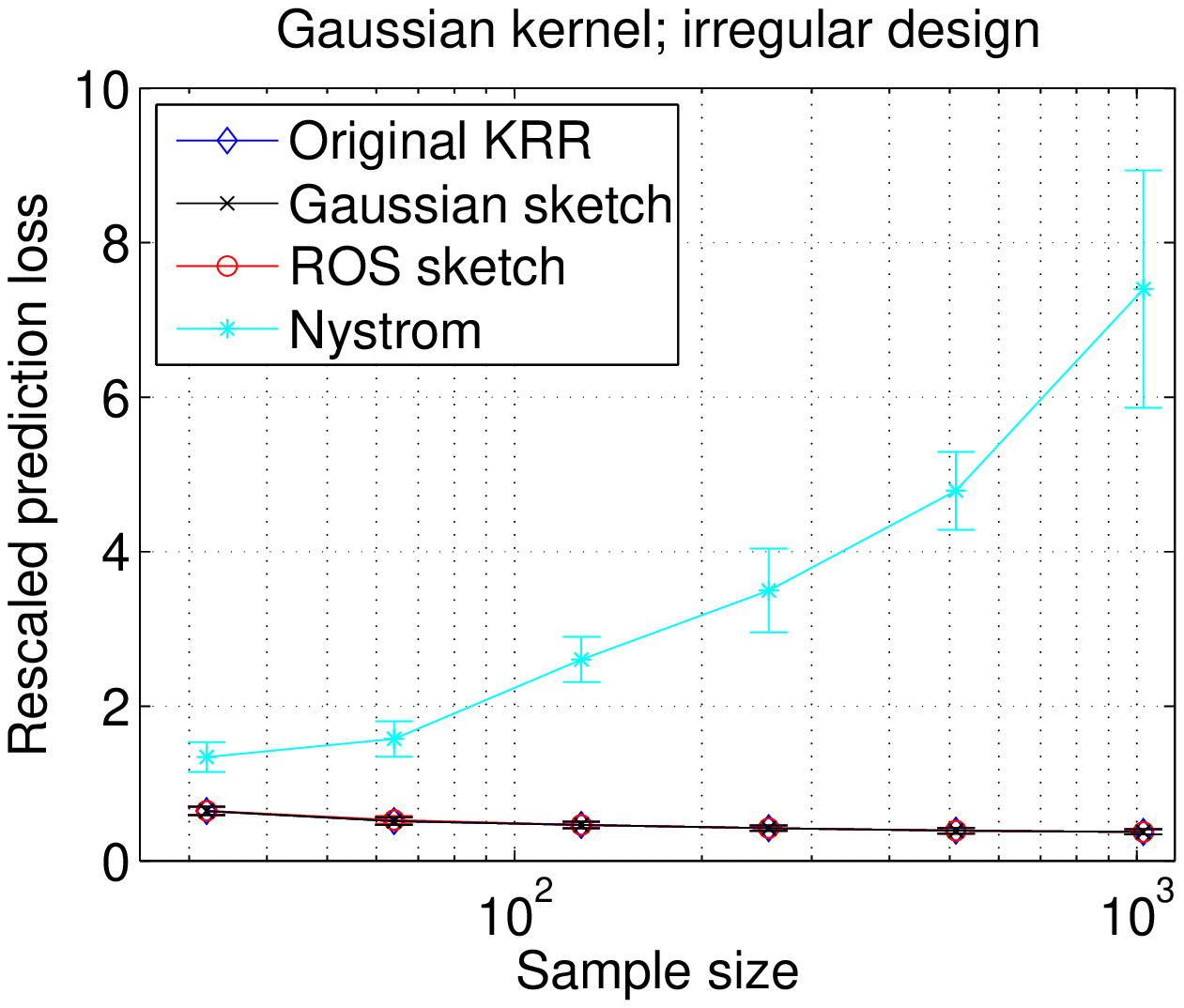} \\
(c) & & (d)
\end{tabular}
\caption{Prediction error versus sample size for original KRR,
  Gaussian sketch, ROS sketch and Nystr\"{o}m approximation.  Left
  panels (a) and (c) shows $\|\fhat-\fstar\|_\numobs^2$ versus the
  sample size $\numobs \in \{32,64,128,256,1024\}$ for projection
  dimension $\numproj=\lceil 4 \sqrt{\log \numobs}\rceil$.  In all
  cases, each point corresponds to the average of $100$ trials, with
  standard errors also shown. Right panels (b) and (d) show the
  rescaled prediction error $\frac{\numobs}{\sqrt{\log \numobs}}
  \|\fhat - \fstar\|_\numobs^2$ versus the sample size.  Top row
  correspond to covariates arranged uniformly on the unit interval,
  whereas bottom row corresponds to an irregular design (see text for
  details).}
\label{FigIrregular}
\end{center}
\end{figure}

The previous example suggests that the \NYS approximation is likely to
be very sensitive to non-inhomogeneity in the sampling of covariates.
In order to explore this conjecture, we performed some additional
simulations, this time comparing both Gaussian and ROS sketches with
the uniform \NYS approximation sketch.  Returning again to the
Gaussian kernel $\KerFunGauss{u}{v} = e^{-\frac{1}{2 h^2} (u -v)^2}$
with bandwidth $h = 0.25$, and the function \mbox{$\fstar(x) = -1 +
  2x^2$,} we first generated $\numobs$ i.i.d. samples that were
uniform on the unit interval $[0,1]$.  We then implemented sketches of
various types (Gaussian, ROS or Nystr\"{o}m) using a sketch dimension
$\numproj = \lceil 4 \sqrt{\log \numobs} \rceil$.  As shown in the top
row (panels (a) and (b)) of Figure~\ref{FigIrregular}, all three
sketch types perform very well for this regular design, with
prediction error that is essentially indistiguishable from the
original KRR estimate.  Keeping the same kernel and function, we then
considered an irregular form of design, namely with $k = \lceil
\sqrt{\numobs} \rceil$ samples perturbed as follows:
\begin{align*}
x_i & \sim \begin{cases} \mbox{Unif}\, [0, 1/2] & \mbox{if $i = 1, \ldots,
    \numobs - k$} \\
1 + z_i & \mbox{for $i = k+1, \ldots, \numobs$}
\end{cases}
\end{align*}
where each $z_i \sim N(0, 1/\numobs)$.  The performance of the
sketched estimators in this case are shown in the bottom row (panels
(c) and (d)) of Figure~\ref{FigIrregular}.  As before, both the
Gaussian and ROS sketches track the performance of the original KRR
estimate very closely; in contrast, the \NYS approximation behaves
very poorly for this regression problem, consistent with the intuition
suggested by the preceding example.

As is known from general theory on the \NYS approximation, its
performance can be improved by knowledge of the so-called leverage
scores of the underlying matrix.  In this vein, recent work by Alaoui
and Mahoney~\cite{Alaoui14} suggests a \NYS approximation non-uniform
sampling of the columns of kernel matrix involving the leverage
scores.  Assuming that the leverage scores are known, they show that
their method matches the performance of original KRR using a
non-uniform sub-sample of the order \mbox{$\trace(\KerMat (\KerMat +
  \lambda_\numobs I)^{-1})\log \numobs)$} columns.  When the
regularization parameter $\lambda_\numobs$ is set optimally---that is,
proportional to $\delcrit^2$---then apart from the extra logarithmic
factor, this sketch size scales with the statistical dimension, as
defined here.  However, the leverage scores are \emph{not known}, and
their method for obtaining a sufficiently approximation requires
sampling $\ptil$ columns of the kernel matrix $\KerMat$, where
\begin{align*}
\ptil \succsim \lambda_\numobs^{-1}\, \trace(\KerMat) \log \numobs.
\end{align*}
For a typical (normalized) kernel matrix $\KerMat$, we have
$\trace(\KerMat) \succsim 1$; moreover, in order to achieve the
minimax rate, the regularization parameter $\regpar_\numobs$ should
scale with $\delcrit^2$.  Putting together the pieces, we see that the
sampling parameter $\ptil$ must satisfy the lower bound $\ptil
\succsim \delcrit^{-2} \log \numobs$.  This requirement is much
larger than the statistical dimension, and prohibitive in many cases:
\begin{itemize}
\item for the Gaussian kernel, we have $\delcrit^2 \asymp
  \frac{\sqrt{\log(\numobs)}}{\numobs}$, and so $\ptil \succsim
  \numobs \log^{1/2}(\numobs)$, meaning that all rows of the kernel
  matrix are sampled.  In contrast, the statistical dimension scales
  as $\sqrt{\log \numobs}$.
\item for the first-order Sobolev kernel, we have $\delcrit^2 \asymp
  \numobs^{-2/3}$, so that $\ptil \succsim \numobs^{2/3} \log
  \numobs$.  In contrast, the statistical dimension for this kernel
  scales as $\numobs^{1/3}$.
\end{itemize}
It remains an open question as to whether a more efficient procedure
for approximating the leverage scores might be devised, which would
allow a method of this type to be statistically optimal in terms
of the sampling dimension.

% the lower bound for $p$ can be
%reduced to $O(\text{tr}(\KerMat (\KerMat + \lambda_\numobs
%I)^{-1})\log n)$, which matches our lower bound for $\numproj$ up to a
%$\log n$ factor. 

%%%%%%%%%%%%%%%%%%%%%%%%%%%%%%%%%%%%%%%%%%%%%%%%%%%%%%%%%%%%%%%%%%%

\section{Proofs}
\label{SecProofs}

In this section, we provide the proofs of our main theorems. Some
technical proofs of the intermediate results are provided in the
appendices.

%%%%%%%%%%%%%%%%%%%%%%%%%%%%%%%%%%%%%%%%%%%%%%%%%%%%%%%%%%%%%%%%%%%%%%%%%%%

\subsection{Proof of Theorem~\ref{ThmMain0}}
\label{SecProofThmMain0}

Recall the definition~\eqref{EqnDefnAlphaDagger} of the estimate
$\fdagger$, as well as the upper bound~\eqref{EqnDecomposition} in
terms of approximation and estimation error terms.  The remainder of
our proof consists of two technical lemmas used to control these two
terms.

\blems[Control of estimation error]
\label{LemEstimationError}
Under the conditions of Theorem~\ref{ThmMain0}, we have
\begin{align}
\label{EqnEstimationError}
\|\fdagger - \fhat\|_\numobs^2 \leq c \: \delcrit^2
\end{align}
with probability at least $1 - c_1 \CEXP{-c_2 \numobs \delcrit^2}$.
\elems

\blems[Control of approximation error]
\label{LemApproxError}
For any $\KerMat$-satisfiable sketch matrix $\Sketch$, we have
\begin{align}
\label{EqnApproxError}
\|\fdagger- \fstar\|_\numobs^2 & \leq \MYHACKCON \; \big \{ \regparn +
\delcrit^2 \big \} \quad \mbox{and}\quad \| \fdagger\|_\Hil \leq
\MYHACKCON \: \Big \{ 1 + \frac{\delcrit^2}{\regparn} \Big \}.
\end{align}

\elems
\vspace*{.05in}

\noindent These two lemmas, in conjunction with the upper
bound~\eqref{EqnDecomposition}, yield the claim in the theorem
statement.  Accordingly, it remains to prove the two lemmas.

%%%%%%%%%%%%%%%%%%%%%%%%%%%%%%%%%%%%%%%%%%%%%%%%%%%%%%%%%%%%%%%%%%%%%%%%%%%%

\subsubsection{Proof of Lemma~\ref{LemEstimationError}}
\label{SecLemEstimationError}

So as to simplify notation, we assume throughout the proof that
$\sigma = 1$.  (A simple rescaling argument can be used to recover the
general statement).  Since $\alphadagger$ is optimal for the quadratic
program~\eqref{EqnDefnAlphaDagger}, it must satisfy the zero gradient
condition
\begin{align}
\label{EqnOptimalCondition}
-\Sketch K \big(\frac{1}{\sqrt{\numobs}} \fstar - \KerMat \Sketch^T
\alphadagger\big) + \regparn \Sketch K \Sketch^T \alphadagger & =0.
\end{align}
By the optimality of $\alphahat$ and feasibility of $\alphadagger$ for
the sketched problem~\eqref{EqnSketchedKRR}, we have
\begin{align*}
\frac{1}{2} \|\KerMat \Sketch^T\alphahat\|_2^2 &-
\frac{1}{\sqrt{\numobs}} y^T \KerMat \Sketch^T\alphahat +
\lambda_\numobs\|\sqrt{\KerMat} \Sketch^T \alphahat\|_2^2 \\
&\leq
\frac{1}{2} \| \KerMat \Sketch^T \alphadagger\|_2^2 -
\frac{1}{\sqrt{\numobs}} y^T \KerMat \Sketch^T \alphadagger+
\lambda_\numobs\|\sqrt{\KerMat} \Sketch^T\alphadagger\|_2^2
\end{align*}
Defining the error vector $\DelHat \defn \Sketch^T(\alphahat -
\alphadagger)$, some algebra leads to the following inequality
\begin{align}
\label{EqnSecondBasicDual}
\frac{1}{2} \|K \DelHat\|_2^2 & \leq -\big\langle K\DelHat, K\Sketch^T
\alphadagger\big\rangle + \frac{1}{\sqrt{n}} y^T K\DelHat +
\lambda_\numobs\|\sqrt{K}\Sketch^T\alphadagger\|_2^2-\lambda_\numobs
\|\sqrt{K}\Sketch^T\alphahat\|_2^2.
\end{align}
Consequently, by plugging in $y = \FstarEval + w$ and applying the
optimality condition~\eqref{EqnOptimalCondition}, we obtain the basic
inequality
\begin{align}
\label{EqnKeyBasic}
\frac{1}{2} \|K \DelHat\|_2^2 & \leq \Big| \frac{1}{\sqrt{n}}w^T
K\DelHat\Big| - \lambda_\numobs\|\sqrt{K}\DelHat\|_2^2.
\end{align}
The following lemma provides control on the right-hand side:
\blems
\label{LemCrossterm}
With probability at least $1- \plaincon_1 e^{-\plaincon_2 \numobs
  \delcrit^2}$, we have
\begin{align}
\label{EqnCrossterm}
\Big| \frac{1}{\sqrt{n}} w^T K \Delta\Big| & \leq
\begin{cases}
6 \delcrit \|K\Delta\|_2 + \HACK \delcrit^2 & \mbox{for all
  $\|\sqrt{K}\Delta\|_2\leq 1$,} \\
2 \delcrit \|K\Delta\|_2 + 2 \delcrit^2 \|\sqrt{K}\Delta\|_2 +
\frac{1}{16} \delcrit^2 & \mbox{for all $\|\sqrt{K}\Delta\|_2 \geq
  1$.}
\end{cases}
\end{align}
\elems

\noindent See Appendix~\ref{AppLemCrossterm} for the proof of this
lemma. \\

\noindent Based on this auxiliary result, we divide the remainder of
our analysis into two cases:
\paragraph{Case 1:} If $\|\sqrt{K}\DelHat\|_2\leq 1$, then the basic 
inequality~\eqref{EqnKeyBasic} and the top inequality in
Lemma~\ref{LemCrossterm} imply
\begin{align}
\label{EqnQuadratic}
\frac{1}{2} \|K \DelHat\|_2^2 & \leq \Big| \frac{1}{\sqrt{n}}w^T
K\DelHat\Big| \leq 6 \delcrit \|K\DelHat\|_2 + \HACK \delcrit^2
\end{align}
with probability at least $1 - \UNICON_1 e^{-\UNICON_2
  n\delcrit^2}$. Note that we have used that fact that the randomness
in the sketch matrix $S$ is independent of the randomness in the
noise vector $w$.  The quadratic inequality~\eqref{EqnQuadratic}
implies that $\|K \DelHat\|_2 \leq c \delcrit$ for some universal
constant $c$.

%%%%%%%%%%%%%%%%%%%%%%%%%%%%%%%%%%%%%%%%%%%%%%%%%%%%%%%%%%%%%%%%%%%%

\paragraph{Case 2:} If $\|\sqrt{\KerMat} \DelHat\|_2>1$, then the basic 
inequality~\eqref{EqnKeyBasic} and the bottom inequality in
Lemma~\ref{LemCrossterm} imply
\begin{align*}
\frac{1}{2} \|K \DelHat\|_2^2 & \leq 2 \delcrit \|K\DelHat\|_2 + 2
\delcrit^2 \|\sqrt{\KerMat} \DelHat\|_2 + \frac{1}{16} \delcrit^2 -
\regparn \|\sqrt{\KerMat} \DelHat\|_2^2
\end{align*}
with probability at least $1- c_1 e^{- c_2 \numobs \delcrit^2}$.  If
$\regparn \geq 2 \delcrit^2$, then under the assumed condition
$\|\sqrt{K}\DelHat\|_2 > 1$, the above inequality gives
\begin{align*}
\frac{1}{2} \|K \DelHat\|_2^2 \leq 2
\delcrit\|K\DelHat\|_2+\frac{1}{16} \delcrit^2 \leq \frac{1}{4} \|K
\DelHat\|_2^2 + 4 \delcrit^2 + \frac{1}{16}\delcrit^2.
\end{align*}
By rearranging terms in the above, we obtain $\|K \DelHat\|_2^2 \leq c
\delcrit^2$ for a universal constant, which completes the proof.

%%%%%%%%%%%%%%%%%%%%%%%%%%%%%%%%%%%%%%%%%%%%%%%%%%%%%%%%%%%%%%%%%%%%%%%%%%%

\subsubsection{Proof of Lemma~\ref{LemApproxError}}
\label{SecProofOfLemApproxError}

Our goal is to show that the bound
\begin{align*}
\frac{1}{2 \numobs} \| \FstarEval -\sqrt{\numobs} \KerMat \Sketch^T
\alphadagger \|_2^2 + \regparn \|\sqrt{\KerMat} \Sketch^T
\alphadagger\|_2^2 \leq \UNICON \big \{ \regparn + \delcrit^2 \big \}.
\end{align*}
In fact, since $\alphadagger$ is a minimizer, it suffices to exhibit
some $\alpha \in \real^\numproj$ for which this inequality holds.
Recalling the eigendecomposition \mbox{$\KerMat = U D U^T$,} it is
equivalent to exhibit some $\alpha \in \real^\numproj$ such that
\begin{align}
\label{EqnGoalOfCon}
\frac{1}{2} \|\thetastar - D \SketchTil^T \alpha\|_2^2 + \regparn
\alpha^T \SketchTil D \SketchTil^T \alpha & \leq c \, \Big \{ \regparn
+ \delcrit^2 \Big \},
\end{align}
where $\SketchTil = \Sketch U$ is the transformed sketch matrix, and
the vector $\thetastar = \numobs ^{-1/2} U \FstarEval \in
\real^\numobs$ satisfies the ellipse constraint $\|D^{-1/2}
\thetastar\|_2 \leq 1$.

We do so via a constructive procedure.  First, we partition the vector
$\thetastar \in \real^\numobs$ into two sub-vectors, namely
$\thetastar_1 \in \real^{\effdim}$ and $\thetastar_2 \in
\real^{\numobs - \effdim}$.  Similarly, we partition the diagonal
matrix $D$ into two blocks, $D_1$ and $D_2$, with dimensions $\effdim$
and $\numobs-\effdim$ respectively.  Under the condition $\numproj >
\effdim$, we may let $\SketchTil_1 \in \real^{\numproj \times
  \effdim}$ denote the left block of the transformed sketch matrix,
and similarly, let $\SketchTil_2 \in \real^{\numproj \times (\numobs -
  \effdim)}$ denote the right block.  In terms of this notation, the
assumption that $\Sketch$ is $\KerMat$-satisfiable corresponds to
the inequalities
\begin{align}
\label{EqnKerSatNew}
\opnorm{\SketchTil_1^T \SketchTil_1 - I_{\statdim}} \leq \frac{1}{2},
\quad \mbox{and} \quad \opnorm{\SketchTil_2 \sqrt{D_2}} \leq c
\delcrit.
\end{align}
As a consequence, we are guarantee that the matrix $\SketchTil_1^T
\SketchTil_1$ is invertible, so that we may define the
$\numproj$-dimensional vector
\begin{align*}
\alphahat & = \SketchTil_1 (\SketchTil_1^T \SketchTil_1)^{-1}
(D_1)^{-1} \betastar_1 \in \real^{\numproj},
\end{align*}
Recalling the disjoint partition of our vectors and matrices, we have
\begin{subequations}
\begin{align}
\label{EqnTermBound}
\|\thetastar - D \SketchTil^T \alphahat\|^2_2 & = \underbrace{\|
  \thetastar_1 - D_1 \SketchTil_1^T \alphahat\|_2}_{=0} +
\underbrace{\|\thetastar_2 - D_2 \SketchTil_2^T \SketchTil_1
  (\SketchTil_1^T \SketchTil_1)^{-1} D_1^{-1} \thetastar_1
  \|^2_2}_{\Term^2_1}
\end{align}
By the triangle inequality, we have
\begin{align*}
\Term_1 & \leq \|\thetastar_2 \|_2 + \|D_2 \SketchTil_2^T \SketchTil_1
(\SketchTil_1^T \SketchTil_1)^{-1} D_1^{-1} \thetastar_1 \|_2 \\
& \leq \|\thetastar_2 \|_2 +\opnorm{D_2 \SketchTil_2^T}
\opnorm{\SketchTil_1} \opnorm{(\SketchTil_1^T \SketchTil_1)^{-1}}
\opnorm{D_1^{-1/2}} \|D_1^{-1/2}\thetastar_1 \|_2 \\
& \leq \|\thetastar_2 \|_2 + \opnorm{\sqrt{D_2}} \opnorm{\SketchTil_2
  \sqrt{D_2}} \opnorm{\SketchTil_1} \opnorm{(\SketchTil_1^T
  \SketchTil_1)^{-1}} \opnorm{D_1^{-1/2}} \|D_1^{-1/2}\thetastar_1
\|_2.
\end{align*}
Since $\|D^{-1/2}
\thetastar\|_2 \leq 1$, we have \mbox{$\|D_1^{-1/2} \thetastar_1\|_2
  \leq 1$} and moreover
\begin{align*}
\|\thetastar_2\|_2^2 = \sum_{j = \effdim+1}^\numobs (\theta^*_j)^2
\leq \delcrit^2 \; \sum_{j = \effdim+1}^\numobs
\frac{(\theta^*_j)^2}{\empkereig_j} \leq \delcrit^2,
\end{align*}
since $\empkereig_j \leq \delcrit^2$ for all $j \geq \effdim + 1$.
Similarly, we have \mbox{$\opnorm{\sqrt{D_2}} \leq
  \sqrt{\empkereig_{\effdim+1}} \leq \delcrit$,} and
\mbox{$\opnorm{D_1^{-1/2}} \leq \delcrit^{-1}$.}  Putting together the pieces,
we have
\begin{align}
\label{EqnBartBound}
\Term_1 & \leq \delcrit + \opnorm{\SketchTil_2 \sqrt{D_2}}
\opnorm{\SketchTil_1} \opnorm{(\SketchTil_1^T \SketchTil_1)^{-1}} \;
\leq \; \big( c \delcrit) \; \sqrt{\frac{3}{2}} \, \; 2 \; = \; c'
\delcrit,
\end{align}
\end{subequations}
where we have invoked the $\KerMat$-satisfiability of the sketch
matrix to guarantee the bounds $\opnorm{\SketchTil_1} \leq
\sqrt{3/2}$, $\opnorm{(\SketchTil_1^T \SketchTil)} \geq 1/2$ and
$\opnorm{\SketchTil_2\, \sqrt{D_2}} \leq c \delcrit$.
Bounds~\eqref{EqnTermBound} and~\eqref{EqnBartBound} in conjunction
guarantee that
\begin{subequations}
\begin{align}
\label{EqnBone}
\|\thetastar - D \SketchTil^T \alphahat\|^2_2 & \leq c \, \delcrit^2,
\end{align}
where the value of the universal constant $c$ may change from line to
line.

Turning to the remaining term on the left-side of
inequality~\eqref{EqnGoalOfCon}, applying the triangle inequality and
the previously stated bounds leads to
\begin{align}
\alphahat^T\SketchTil D \SketchTil^T\alphahat & \leq \| D_1^{-1/2}
\thetastar_1\|_2^2 + \opnorm{D_2^{1/2} \SketchTil_2^T}
\opnorm{\SketchTil_1} \nonumber\\
&\qquad\cdot\opnorm{(\SketchTil_1^T \SketchTil_1)^{-1}}
\opnorm{D_1^{-1/2}} \|D_1^{-1/2}\thetastar_1 \|_2 \nonumber\\
\label{EqnBtwo}
& \leq 1 + \big( c \delcrit \big) \; \sqrt{3/2} \; \frac{1}{2} \;
\delcrit^{-1} \, \big (1 \big) 
\leq \; c'.
\end{align}
\end{subequations}
Combining the two bounds~\eqref{EqnBone} and~\eqref{EqnBtwo} yields
the claim~\eqref{EqnGoalOfCon}.

%%%%%%%%%%%%%%%%%%%%%%%%%%%%%%%%%%%%%%%%%%%%%%%%%%%%%%%%%%%%%%%%%%%%%%

\section{Discussion}
\label{SecDiscussion}

In this paper, we have analyzed randomized sketching methods for
kernel ridge regression.  Our main theorem gives sufficient conditions
on any sketch matrix for the sketched estimate to achieve the minimax
risk for non-parametric regression over the underlying kernel class.
We specialized this general result to two broad classes of sketches,
namely those based on Gaussian random matrices and randomized
orthogonal systems (ROS), for which we proved that a sketch size
proportional to the statistical dimension is sufficient to achieve the
minimax risk.  More broadly, we suspect that sketching methods of the
type analyzed here have the potential to save time and space in other
forms of statistical computation, and we hope that the results given
here are useful for such explorations.

%%%%%%%%%%%%%%%%%%%%%%%%%%%%%%%%%%%%%%%%%%%%%%%%%%%%%%%%%%%%%%%%%%%%%%%%%

\subsection*{Acknowledgements}
All authors were partially supported by Office of Naval Research MURI
grant N00014-11-1-0688, National Science Foundation Grants
CIF-31712-23800 and DMS-1107000, and Air Force Office of Scientific
Research grant AFOSR-FA9550-14-1-0016.  In addition, MP was supported
by a Microsoft Research Fellowship.

%% APPENDIX %%%%%%%%%%%%%%%%%%%%%%%%%%%%%%%%%%%%%%%%%%%%%%%%%%%%%%%%%%%%

\appendix

\section{Subsampling sketches yield \NYS approximation}
\label{AppNystrom}

In this appendix, we show that the the sub-sampling sketch matrix
described at the end of Section~\ref{SecKernelSketchBackground}
coincides with applying Nystr{\"o}m
approximation~\citep{williams2001Nystrom} to the kernel matrix.

We begin by observing that the original KRR quadratic
program~\eqref{EqnOriginalKRR} can be written in the equivalent form
$\min \limits_{\weight \in \real^\numobs,\, u \in \real^\numobs } \{
\frac{1}{2 \numobs} \|u\|^2 + \regparn \weight^T \KerMat \weight \}$
such that $y-\sqrt{n}\KerMat \weight = u$.  The dual of this
constrained quadratic program (QP) is given by
\begin{align}
\label{EqnDualProblem}
\xi^\dagger & = \arg \max_{\xi \in\real^n} \Big \{-
\frac{n}{4\lambda_n} \xi^T K \xi + \xi^T y - \frac{1}{2}\xi^T \xi \Big
\}.
\end{align}
The KRR estimate $\fdagger$ and the original solution $\wdagger$ can
be recovered from the dual solution $\xi^\dagger$ via the relation
$\fdagger(\cdot) = \frac{1}{\sqrt{\numobs}} \sum_{i=1}^\numobs
\wdagger_i \KerFun{\cdot}{x_i}$ and
$\wdagger=\frac{\sqrt{n}}{2\lambda_n} \xi^\dagger$.
 
Now turning to the the sketched KRR program~\eqref{EqnSketchedKRR},
note that it can be written in the equivalent form \mbox{$\min
  \limits_{\alpha \in \real^\numobs,\, u \in \real^\numobs } \big \{
  \frac{1}{2 \numobs} \|u\|^2 + \regparn \alpha^T \Sketch\KerMat
  \Sketch^T\alpha \big \}$} subject to the constraint
\mbox{$y-\sqrt{n}\KerMat \Sketch^T \alpha = u$.}  The dual of this
constrained QP is given by
\begin{align}
\label{EqnDualSketchProblem}
\xi^\ddagger & = \arg \max_{\xi \in\real^n} \Big \{-
\frac{n}{4\lambda_n} \xi^T \widetilde K \xi + \xi^T y -
\frac{1}{2}\xi^T \xi \Big \},
\end{align}
where $\widetilde K =\KerMat \Sketch^T(\Sketch
\KerMat\Sketch^T)^{-1}\Sketch\KerMat$ is a rank-$\numproj$ matrix in
$\real^{\numobs\times\numobs}$.  In addition, the sketched KRR
estimate $\fhat$, the original solution $\alphahat$ and the dual
solution $\xi^\ddagger$ are related by
\mbox{$\fhat(\cdot)=\frac{1}{\sqrt{\numobs}} \sum_{i=1}^\numobs
  (\Sketch^T \alphahat)_i \KerFun{\cdot}{x_i}$} and \mbox{$\alphahat =
  \frac{\sqrt{n}}{2\lambda_n}(\Sketch
  \KerMat\Sketch^T)^{-1}\Sketch\KerMat \xi^\ddagger$.}

When $\Sketch$ is the sub-sampling sketch matrix, the matrix
$\widetilde K = \KerMat \Sketch^T (\Sketch \KerMat \Sketch^T)^{-1}$ $
\Sketch \KerMat$ is known as the \NYS
approximation~\citep{williams2001Nystrom}. Consequently, the dual
formulation of sketched KRR based on a sub-sampling matrix can be
viewed as the \NYS approximation as applied to the dual formulation of
the original KRR problem.

%%%%%%%%%%%%%%%%%%%%%%%%%%%%%%%%%%%%%%%%%%%%%%%%%%%%%%%%%%%%%%%%%%%%%%

\section{Technical Proofs}

\subsection{Proof of Theorem~\ref{ThmLowerBound}}
\label{SecLowerBound}
We begin by converting the problem to an instance of the normal
sequence model~\cite{Johnstone12}.  Recall that the kernel matrix can
be decomposed as $\KerMat = U^T D U$, where $U \in \real^{\numobs
  \times \numobs}$ is orthonormal, and $D = \diag \{ \empkereig_1,
\ldots, \empkereig_\numobs \}$.  Any function $\fstar \in \Hil$ can be
decomposed as
\begin{align}
\label{EqnFstarRep}
\fstar & = \frac{1}{\sqrt{\numobs}} \sum_{j=1}^\numobs
\KerFun{\cdot}{x_j} (U^T \betastar)_j + g,
\end{align}
for some vector $\betastar \in \real^\numobs$, and some function $g
\in \Hil$ is orthogonal to $\myspan\{$ $\KerFun{\cdot}{x_j}, j = 1,
\ldots, \numobs \}$.  Consequently, the inequality $\|\fstar\|_\Hil
\leq 1$ implies that
\begin{align*}
\Big \| \frac{1}{\sqrt{\numobs}} \sum_{j=1}^\numobs
\KerFun{\cdot}{x_j} (U^T \betastar)_j \Big \|_\Hil^2 \; = \; \big (U^T
\betastar \big)^T U^T D U \big(U^T \betastar \big) \; = \; \|\sqrt{D}
\betastar\|_2^2 \leq 1.
\end{align*}
Moreover, we have $\fstar(x_1^\numobs) = \sqrt{\numobs} U^T D
\betastar$, and so the original observation model~\eqref{EqnModel} has 
the equivalent form $y = \sqrt{\numobs} U^T \thetastar + w$, where
$\thetastar = D \betastar$.  In fact, due to the rotation invariance
of the Gaussian, it is equivalent to consider the normal sequence
model
\begin{align}
\label{EqnSimple}
\ytil = \thetastar + \frac{w}{\sqrt{\numobs}}.
\end{align}
Any estimate $\thetatil$ of $\thetastar$
defines the function estimate $\ftil(\cdot) = \frac{1}{\sqrt{\numobs}}
\sum_{i=1}^\numobs \KerFun{\cdot}{x_i} $ $\big(U^T D^{-1} \thetatil)_i$,
and by construction, we have $\|\ftil - \fstar\|_\numobs^2 =
\|\thetatil - \thetastar\|_2^2$.  Finally, the original constraint $
\|\sqrt{D} \betastar\|_2^2 \leq 1$ is equivalent to $\|D^{-1/2}
\thetastar\|_2 \leq 1$.  Thus, we have a version of the normal
sequence model subject to an ellipse constraint. \\

After this reduction, we can assume that we are given $\numobs$ i.i.d.
observations $\ytil_1^\numobs = \{\ytil_1, \ldots, \ytil_\numobs \}$,
and our goal is to lower bound the Euclidean error $\|\thetatil -
\thetastar\|_2^2$ of any estimate of $\thetastar$.  In order to do so,
we first construct a $\delta/2$-packing of the set $\BasicEllipse = \{
\theta \in \real^\numobs \, \mid \, \|D^{-1/2} \theta\|_2 \leq 1 \}$,
say $\{\theta^1, \ldots, \ldots, \theta^M\}$.  Now consider the random
ensemble of regression problems in which we first draw an index
$\RandInd$ uniformly at random from the index set $[M]$, and then
conditioned on $\RandInd = \randind$, we observe $\numobs$
i.i.d. samples from the non-parametric regression model with $\fstar =
f^\randind$.  Given this set-up, a standard argument using Fano's
inequality implies that
\begin{align*}
\mprob \big[ \|\flest - \fstar\|_\numobs^2 \geq \frac{\delta^2}{4}
  \big] & \geq 1 - \frac{I(\ytil_1^\numobs; \RandInd) + \log 2}{\log
  M},
\end{align*}
where $I(\ytil_1^\numobs; \RandInd)$ is the mutual information between
the samples $\ytil_1^\numobs$ and the random index $\RandInd$.  It
remains to construct the desired packing and to upper bound the mutual
information.

For a given $\delta > 0$, define the ellipse
\begin{align}
\label{EqnDefnEllipse}
\Ellipse(\delta) & \defn \Big \{ \theta \in \real^\numobs \, \mid \,
\underbrace{\sum_{j=1}^\numobs
  \frac{\theta_j^2}{\hackmu{j}}}_{\ELLNORM{\theta}^2} \leq 1 \Big \}.
\end{align}
By construction, observe that $\Ellipse(\delta)$ is contained within
Hilbert ball of unit radius.  Consequently, it suffices to construct a
$\delta/2$-packing of this ellipse in the Euclidean norm.

\blems
\label{LemPackingEllipse}
For any $\delta \in(0,\delcrit]$, there is a $\delta/2$-packing of the ellipse
$\Ellipse(\delta)$ with cardinality
\begin{align}
\log M & = \frac{1}{64} \statdim.
\end{align}

\elems

Taking this packing as given, note that by construction, we have
\begin{align*}
\|\theta^\randind\|_2^2 = \delta^2 \sum_{j=1}^\numobs
\frac{(\theta^\randind)_j^2}{\delta^2} \leq \delta^2, \quad \mbox{and
  hence} \quad \|\theta^\randind - \theta^{\randindb}\|_2^2 \leq 4
\delta^2.
\end{align*}
In conjunction with concavity of the KL diveregence, we have
\begin{align*}
I(y_1^\numobs; J) & \leq \frac{1}{M^2} \sum_{\randind, \randindb=1}^M
\kull{\mprob^\randind}{\mprob^\randindb} = \frac{1}{M^2}
\frac{\numobs}{2 \sigma^2} \sum_{\randind, \randindb=1}^M
\|\theta^\randind - \theta^\randindb\|_2^2 \; \leq \;
\frac{2\numobs}{\sigma^2} \delta^2
\end{align*} 
For any $\delta$ such that $\log 2 \leq \frac{2 \numobs}{\sigma^2}
\delta^2$ and $\delta\leq \delcrit$, we have
\begin{align*}
\mprob \Big[ \|\flest - \fstar\|_\numobs^2 \geq \frac{\delta^2}{4}
  \Big] & \geq 1 - \frac{ 4 \numobs
  \delta^2/\sigma^2}{\statdim/64}.
\end{align*}
Moreover, since the kernel is regular, we have $\sigma^2 \statdim\geq \UNICON \numobs\delcrit^2$ for some positive constant $\UNICON$.
 Thus, setting $\delta^2 = 
\frac{\UNICON\delcrit^2}{512}$ yields the claim.

\paragraph{Proof of Lemma~\ref{LemPackingEllipse}:}

It remains to prove the lemma, and we do so via the probabilistic
method.  Consider a random vector $\theta \in \real^\numobs$ of the
form
\begin{align}
\label{EqnOriginalEnsemble}
\theta & = \begin{bmatrix} \frac{\delta}{\sqrt{2 \statdim}} w_1 &
  \frac{\delta}{\sqrt{2 \statdim}} w_2 & \cdots &
  \frac{\delta}{\sqrt{2 \statdim}} w_\statdim & 0 & \cdots & 0
\end{bmatrix},
\end{align}
where $w=(w_1,\ldots,w_\statdim)^T \sim N(0, I_\statdim)$ is a standard Gaussian vector.  We
claim that a collection of $M$ such random vectors $\{\theta^1,
\ldots, \theta^M \}$, generated in an i.i.d.  manner, defines the
required packing with high probability.

On one hand, for each index $\randind \in [M]$, since $\delta^2\leq\delcrit^2\leq \empkereig_j$ for each $j\leq \statdim$, we have
$\|\theta^\randind\|_\Ellipse^2 = \frac{\|w^\randind\|_2^2}{2\statdim}$, corresponding to a normalized $\chi^2$-variate. 
Consequently, by a combination of standard tail
bounds and the union bound, we have
\begin{align*}
\mprob \Big[ \|\theta^\randind\|_\Ellipse^2 \leq 1 \quad \mbox{for all
    $\randind \in [M]$} \Big] & \geq 1-M \, e^{- \frac{\statdim}{16}}.
\end{align*}

Now consider the difference vector $\theta^\randind -
\theta^\randindb$.  Since the underlying Gaussian noise vectors
$w^\randind$ and $w^\randindb$ are independent, the difference vector
$w^\randind - w^\randindb$ follows a $N(0, 2 I_\numproj)$ distribution.
Consequently, the event $\|\theta^\randind - \theta^\randindb\|_2 \geq
\frac{\delta}{2}$ is equivalent to the event $\sqrt{2}\|\theta\|_2
\geq \frac{\delta}{2}$, where $\theta$ is a random vector drawn from
the original ensemble. Note that $\|\theta\|_2^2=\delta^2\frac{\|w\|_2^2}{2\statdim}$. Then  a combination of standard tail
bounds for $\chi^2$-distributions and the union bound argument yields
\begin{align*}
\mprob \Big[ \|\theta^\randind-\theta^\randindb\|_2^2 \geq \frac{\delta^2}{4} \quad \mbox{for all
    $\randind,\, \randindb \in [M]$} \Big] & \geq 1-M^2 \, e^{- \frac{\statdim}{16}}.
\end{align*}

Combining the last two display together, we obtain
\begin{align*}
&\mprob \Big[  \|\theta^\randind\|_\Ellipse^2 \leq 1\mbox{ and }\|\theta^\randind-\theta^\randindb\|_2^2 \geq \frac{\delta^2}{4} \quad \mbox{for all
    $\randind,\, \randindb \in [M]$} \Big] \\
  \geq&\, 1-M \, e^{- \frac{\statdim}{16}}-M^2 \, e^{- \frac{\statdim}{16}}.
\end{align*}
This probability is positive for $\log M = \statdim/64$.

%%%%%%%%%%%%%%%%%%%%%%%%%%%%%%%%%%%%%%%%%%%%%%%%%%%%%%%%%%%%%%%%%%%%%%%%%%

\subsection{Proof of Lemma~\ref{LemCrossterm}}
\label{AppLemCrossterm}

For use in the proof, for each $\delta > 0$, let us define the random
variable
\begin{align}
\label{EqnHanaChan}
Z_\numobs(\delta) & = \sup_{ \substack{\|\sqrt{K} \Delta \|_2 \leq 1
    \\ \|K \Delta\|_2 \leq \delta}} \Big| \frac{1}{\sqrt{\numobs}} w^T
K \Delta \Big|.
\end{align}

\paragraph{Top inequality in the bound~\eqref{EqnCrossterm}:} If the 
top inequality is violated, then we claim that we must have
$Z_\numobs(\delcrit) > \HACK \delcrit^2$.  On one hand, if the
bound~\eqref{EqnCrossterm} is violated by some vector $\Delta \in
\real^\numobs$ with $\|K \Delta\|_2 \leq \delcrit$, then we have
\begin{align*}
\HACK \delcrit^2 \; \leq \; \Big| \frac{1}{\sqrt{n}} w^T K \Delta\Big|
& \leq Z_\numobs(\delcrit).
\end{align*}
On the other hand, if the bound is violated by some function with $\|K
\Delta\|_2 > \delcrit$, then we can define the rescaled vector
$\DelTil = \frac{\delcrit}{\|K \Delta\|_2} \, \Delta$, for which we
have
\begin{align*}
\|K \DelTil\|_2 = \delcrit, \quad \mbox{and} \quad \| \sqrt{K}
\DelTil\|_2 = \frac{\delcrit}{\|K \Delta\|_2} \|\sqrt{K} \Delta\|_2 \;
\leq \; 1
\end{align*}
showing that $Z_\numobs(\delcrit) \geq \HACK \delcrit^2$ as well.

When viewed as a function of the standard Gaussian vector $w \in
\real^\numobs$, it is easy to see that $Z_\numobs(\delcrit)$ is
Lipschitz with parameter $\delcrit/\sqrt{\numobs}$.  Consequently, by
concentration of measure for Lipschitz functions of
Gaussians~\cite{Ledoux01}, we have
\begin{align}
\label{EqnLedouxTail}
\mprob \big[ Z_\numobs(\delcrit) \geq \Exs[Z_\numobs(\delcrit)] + t \big]
& \leq e^{- \frac{\numobs t^2}{2 \delcrit^2}}.
\end{align}
Moreover, we claim that
\begin{align}
\label{EqnCoffee}
\Exs[Z_\numobs(\delcrit)] & \stackrel{(i)}{\leq}
\underbrace{\sqrt{\frac{1}{\numobs} \sum_{i=1}^\numobs \min \{
    \delcrit^2, \empkereig_j \}}}_{\EmpKerComp(\delcrit)} \;
\stackrel{(ii)}{\leq} \; \delcrit^2
\end{align}
where inequality (ii) follows by definition of the critical radius
(recalling that we have set $\sigma = 1$ by a rescaling argument).
Setting $t = \delcrit^2$ in the tail bound~\eqref{EqnLedouxTail}, we
see that $\mprob[Z_\numobs(\delcrit) \geq \HACK \delcrit^2] \leq
e^{\numobs \delcrit^2/2}$, which completes the proof of the top bound.

It only remains to prove inequality (i) in equation~\eqref{EqnCoffee}.
The kernel matrix $K$ can be decomposed as $K = U^T D U$, where $D =
\diag \{ \empkereig_1, \ldots, \empkereig_\numobs \}$, and $U$ is a
unitary matrix.  Defining the vector $\beta = D U \Delta$, the two
constraints on $\Delta$ can be expressed as $\|D^{-1/2} \beta\|_2 \leq
1$ and $\|\beta\|_2 \leq \delta$.  Note that any vector satisfying
these two constraints must belong to the ellipse
\begin{align*}
\Ellipse & \defn \Big \{ \beta \in \real^\numobs \, \mid \,
\sum_{j=1}^\numobs \frac{\beta_j^2}{\nu_j} \leq 2 \qquad \mbox{where
  $\nu_j = \max \{ \delcrit^2, \empkereig_j \}$} \Big \}.
\end{align*}
Consequently, we have
\begin{align*}
\Exs[Z_\numobs(\delcrit)] & \leq \Exs \Big[\sup_{\beta \in \Ellipse}
  \frac{1}{\sqrt{\numobs}} \big| \inprod{U^T w}{\beta} \big| \Big] \;
= \; \Exs \Big[\sup_{\beta \in \Ellipse} \frac{1}{\sqrt{\numobs}}
  \big| \inprod{w}{\beta} \big| \Big],
\end{align*}
since $U^T w$ also follows a standard normal distribution.  By the
Cauchy-Schwarz inequality, we have
\begin{align*}
\Exs \Big[\sup_{\beta \in \Ellipse} \frac{1}{\sqrt{\numobs}} \big|
  \inprod{w}{\beta} \big| \Big] & \leq \frac{1}{\sqrt{\numobs}} \Exs
\sqrt{\sum_{j=1}^\numobs \nu_j w_j^2} \; \leq \;
\underbrace{\frac{1}{\sqrt{\numobs}} \sqrt{\sum_{j=1}^\numobs \nu_j
}}_{\EmpKerComp(\delcrit)},
\end{align*}
where the final step follows from Jensen's inequality.

\paragraph{Bottom inequality in the bound~\eqref{EqnCrossterm}:}
We now turn to the proof of the bottom inequality.  We claim that it
suffices to show that
\begin{align}
\label{EqnSuffice}
\Big|\frac{1}{\sqrt{\numobs}} w^T K \DelTil \Big| & \leq 2 \, \delcrit
\|K \DelTil\|_2 + 2 \, \delcrit^2 + \TWOHACK \|K \DelTil\|_2^2
\end{align}
for all $\DelTil \in \real^\numobs$ such that $\|\sqrt{K}
  \DelTil\|_2 = 1$.
Indeed, for any vector $\Delta \in \real^\numobs$ with $\|\sqrt{K}
\Delta\|_2 > 1$, we can define the rescaled vector $\DelTil =
\Delta/\|\sqrt{K} \Delta\|_2$, for which we have $\|\sqrt{K}
\DelTil\|_2 = 1$.  Applying the bound~\eqref{EqnSuffice} to this
choice and then multiplying both sides by $\|\sqrt{K} \Delta\|_2$, we
obtain
\begin{align*}
\Big|\frac{1}{\sqrt{\numobs}} w^T K \Delta \Big| & \leq 2 \, \delcrit
\|K \Delta\|_2 + 2 \, \delcrit^2 \| \sqrt{K} \Delta\|_2 + \TWOHACK
\frac{\|K \Delta\|_2^2}{\|\sqrt{K} \Delta\|_2} \\
& \leq 2 \, \delcrit \|K \Delta\|_2 + 2 \, \delcrit^2 \| \sqrt{K}
\Delta\|_2 + \TWOHACK \|K \Delta\|_2^2,
\end{align*}
as required.

Recall the family of random variables $Z_\numobs$ previously
defined~\eqref{EqnHanaChan}.  For any $u \geq \delcrit$, we have
\begin{align*}
\Exs[Z_\numobs(u)] & = \EmpKerComp(u) \; = \; u
\frac{\EmpKerComp(u)}{u} \; \stackrel{(i)}{\leq} \; u \frac{
  \EmpKerComp(\delcrit)}{\delcrit} \; \stackrel{(ii)}{\leq} \; u
\delcrit,
\end{align*}
where inequality (i) follows since the function $u \mapsto
\frac{\EmpKerComp(u)}{u}$ is non-increasing, and step (ii) follows by
our choice of $\delcrit$.  Setting $t = \frac{u^2}{32}$ in the
concentration bound~\eqref{EqnLedouxTail}, we conclude that
\begin{align}
\label{EqnChocolateBound}
\mprob \big[Z_\numobs(u) \geq u \delcrit + \frac{u^2}{64} \big] & \leq
\CEXP{ - \plaincon \numobs u^2 } \quad \mbox{for each $u \geq
  \delcrit$.}
\end{align}

We are now equipped to prove the bound~\eqref{EqnSuffice} via a
``peeling'' argument.  Let $\Event$ denote the event that the
bound~\eqref{EqnSuffice} is violated for some vector $\DelTil$ with
$\|\sqrt{K} \DelTil\|_2 = 1$.  For real numbers $0 \leq a < b$, let
$\Event(a,b)$ denote the event that it is violated for some vector
with $\|\sqrt{K} \Delta\|_2 = 1$ and $\|K \DelTil\|_2 \in [a,b]$.  For
$m = 0, 1, 2, \ldots$, define $u_m = 2^{m} \delcrit$. We then have the
decomposition $\Event = \Event(0, u_0) \cup \big( \bigcup_{m=0}^\infty
\Event(u_m, u_{m+1}) \big)$ and hence by union bound,
\begin{align}
\label{EqnEarlyUnion}
\mprob[\Event] & \leq \mprob[\Event(0,u_0)] + \sum_{m=0}^\infty
\mprob[\Event(u_m, u_{m+1})].
\end{align}
The final step is to bound each of the terms in this summation, Since
$u_0 = \delcrit$, we have
\begin{align}
\label{EqnZeroBound}
\mprob[\Event(0, u_0)] & \leq \mprob[Z_\numobs(\delcrit) \geq 2
  \delcrit^2] \; \leq \; \CEXP{ - \plaincon \numobs \delcrit^2}.
\end{align}
On the other hand, suppose that $\Event(u_m, u_{m+1})$ holds, meaning
that there exists some vector $\DelTil$ with $\|\sqrt{K}\DelTil\|_2 =
1$ and $\|K \DelTil\|_2 \in [u_m, u_{m+1}]$ such that
\begin{align*}
\Big|\frac{1}{\sqrt{\numobs}} w^T K \DelTil \Big| & \geq 2 \, \delcrit
\|K \DelTil\|_2 + 2 \, \delcrit^2 + \TWOHACK \|K \DelTil\|_2^2 \\
& \geq 2 \delcrit u_m + 2 \delcrit^2 + \frac{1}{16} u_m^2 \\
& \geq \delcrit u_{m+1} + \frac{1}{64} u_{m+1}^2,
\end{align*}
where the second inequality follows since $\|K \DelTil\|_2 \geq u_m$;
and the third inequality follows since $u_{m+1} = 2 u_m$.  This lower
bound implies that $Z_\numobs(u_{m+1}) \geq \delcrit u_{m+1} +
\frac{u_{m+1}^2}{64}$, whence the bound~\eqref{EqnChocolateBound}
implies that
\begin{align*}
\mprob \big[\Event(u_m, u_{m+1})] & \leq \CEXP{ - \plaincon \numobs
  u_{m+1}^2} \leq \CEXP{ - \plaincon \numobs \, 2^{2m} \delcrit^2}.
\end{align*}
Combining this tail bound with our earlier bound~\eqref{EqnZeroBound}
and substituting into the union bound~\eqref{EqnEarlyUnion} yields
\begin{align*}
\mprob[\Event] & \leq \CEXP{ -\plaincon \numobs \delcrit^2} +
\sum_{m=0}^\infty \exp \big( - \plaincon \numobs \, 2^{2m} \delcrit^2
\big) \; \leq \; \plaincon_1 \CEXP{- \plaincon_2 \numobs \delcrit^2},
\end{align*}
as claimed.

%%%%%%%%%%%%%%%%%%%%%%%%%%%%%%%%%%%%%%%%%%%%%%%%%%%%%%%%%%%%%%%%%%%%%%

\subsection{Proof of Corollary~\ref{CorSubgaussian}}
\label{SecProofCorSubgaussian}

Based on Theorem~\ref{ThmMain0}, we need to verify that the stated
lower bound~\eqref{EqnDefnKgood} on the projection dimension is
sufficient to guarantee that that a random sketch matrix is
$\KerMat$-satisfiable is high probability.  In particular, let us
state this guarantee as a formal claim:

\blems
\label{LemRandMat}
Under the lower bound~\eqref{EqnDefnKgood} on the sketch
dimension, a $\{$Gaussian, ROS$\}$ random sketch is
$\KerMat$-satisfiable with probability at least $\PHIPROB$.
\elems

\noindent We split our proof into two parts, one for each inequality
in the definition~\eqref{EqnKerSat} of $\KerMat$-satisfiability.

\subsubsection{Proof of inequality (i):}

We need to bound the operator norm of the matrix $Q = U_1^T \Sketch^T
\Sketch U_1- I_{\effdim}$, where the matrix $U_1 \in \real^{\numobs
  \times \effdim}$ has orthonormal columns. Let $\{v^1, \ldots, v^N
\}$ be a $1/2$-cover of the Euclidean sphere $\Sphere{\effdim}$; by
standard arguments~\cite{Matousek}, we can find such a set with $N
\leq \CEXP{2 \effdim}$ elements.  Using this cover, a straightforward
discretization argument yields
\begin{align*}
\opnorm{Q} & \leq 4 \max_{j,k=1, \ldots, N} \inprod{v^j}{Q v^k} \; =
\; 4 \max_{j, k = 1, \ldots, N} (\vtil)^j \Big \{ \Sketch^T \Sketch -
I_{\numobs} \Big \} \vtil^k,
\end{align*}
where $\vtil^j \defn U_1 v^j \in \Sphere{\numobs}$, and $\Qtil =
\Sketch^T \Sketch - I_\numobs$.  In the Gaussian case, standard
sub-exponential bounds imply that $\mprob\big[(\vtil)^j \Qtil \vtil^k
  \geq 1/8 \big] \leq c_1 \CEXP{-c_2 \numproj}$, and consequently, by
the union bound, we have
\begin{align*}
\mprob \big[\opnorm{Q} \geq 1/2] \leq c_1 \CEXP{-c_2 \numproj + 4
  \effdim } \; \leq c_1 \CEXP{-c'_2 \numproj},
\end{align*}
where the second and third steps uses the assumed lower bound on
$\numproj$.  In the ROS case, results of Krahmer and
Ward~\cite{KraWar11} imply that
\begin{align*}
\mprob \big[\opnorm{Q} \geq 1/2] & \leq c_1 \CEXP{-c_2
  \frac{\numproj}{\log^4(\numobs)}}.
\end{align*}
where the final step uses the assumed lower bound on $\numproj$.

\subsubsection{Proof of inequality (ii):}  We split this claim into two
sub-parts: one for Gaussian sketches, and the other for ROS sketches.
Throughout the proof, we make use of the $\numobs \times \numobs$
diagonal matrix \mbox{$\Dbar = \diag(0_{\effdim}, D_2)$,} with which
we have $\Sketch U_2 D_2^{1/2} = \Sketch U \Dbar^{1/2}$.

\paragraph{Gaussian case:}

By the definition of the matrix spectral norm, we know
\begin{align}
\label{EqnZvar}
\opnorm{\Sketch U \Dbar^{1/2}} & \defn \sup_{ \substack{ u \in
    \Sphere{\numproj} \\ v \in \Ellipse }} \inprod{u}{\Sketch v},
\end{align}
where $\Ellipse = \{v \in \real^\numobs \, \mid \, \|U \Dbar v\|_2
\leq 1 \}$, and $\Sphere{\numproj} = \{ u \in \real^\numproj \, \mid
\, \|u\|_2 = 1 \}$.

We may choose a $1/2$-cover $\{u^1, \ldots, u^M \}$ of the set
$\Sphere{\numproj}$ of the set with $\log M \leq 2 \numproj$
elements. We then have
\begin{align*}
\opnorm{\Sketch U \Dbar^{1/2}} & \leq \max_{j \in [M]} \sup_{v \in
  \Ellipse} \inprod{u^j}{\Sketch v} + \frac{1}{2} \sup_{ \substack{ u
    \in \Sphere{\effdim} \\ v \in \Ellipse }} \inprod{u}{\Sketch v} \\
    &=
\max_{j \in [M]} \sup_{v \in \Ellipse} \inprod{u^j}{\Sketch v} +
\frac{1}{2} \opnorm{\Sketch U \Dbar^{1/2}},
 \end{align*}
and re-arranging implies that
\begin{align*}
\opnorm{\Sketch U \Dbar^{1/2}} & \leq 2 \underbrace{\max_{j \in [M]}
  \sup_{v \in \Ellipse} \inprod{u^j}{ \SketchTil v}}_{\Ztil}.
\end{align*}
For each fixed $u^j \in \Sphere{\effdim}$, consider the random
variable $Z^j \defn \sup_{v \in \Ellipse} \inprod{u^j}{\Sketch v}$.
It is equal in distribution to the random variable
$V(g)=\frac{1}{\sqrt{\numproj}} \sup_{v \in\Ellipse} \inprod{g}{v}$,
where $g \in \real^{\numobs}$ is a standard Gaussian vector.  For $g,
g'\in \real^{\numobs}$, we have
\begin{align*}
|V(g) - V(g')| & \leq \frac{2}{\sqrt{\numproj}} \sup_{v \in \Ellipse}
|\inprod{g - g'}{v} | \\
&\leq 
\frac{2\opnorm{D_2^{1/2}}}{\sqrt{\numproj}} \|g - g'\|_2 \; \leq \;
\frac{2 \delcrit}{\sqrt{\numproj}} \|g - g'\|_2,
\end{align*}
where we have used the fact that $\empkereig_j \leq \delcrit^2$ for
all $j \geq \effdim+1$. Consequently, by concentration of measure for
Lipschitz functions of Gaussian random variables \cite{Ledoux01}, we
have
\begin{align}
\label{EqnLedoux}
\mprob \big[ V(g)\geq \Exs[V(g)] + t \big] \leq e^{- \frac{\numproj
    t^2}{8 \delcrit^2}}.
\end{align}
Turning to the expectation, we have
\begin{align}
\Exs[V(g)] &=\frac{2}{\sqrt{\numproj}} \Exs \big\|D_2^{1/2}g\big\|_2
\; \leq 2 \sqrt{ \frac{ \sum_{j = \effdim +1}^\numobs
    \mu_j}{\numproj}} \; = 2 \sqrt{ \frac{\numobs}{\numproj} }
\sqrt{\frac{ \sum_{j = \effdim+1}^\numobs \mu_j}{\numobs}} \leq 2
\delcrit
\end{align}
 where the last inequality follows since $\numproj \geq \numobs
 \delcrit^2$ and $\sqrt{\frac{ \sum_{j = \effdim+1}^\numobs
     \mu_j}{\numobs}} \leq \delcrit^2$.  Combining the pieces, we have
 shown have shown that $\mprob[Z^j\geq c_0 (1+\epsilon) \delcrit] \leq
 e^{ -c_2 \numproj}$ for each $j=1,\ldots,M$.  Finally, setting $t = c
 \delcrit$ in the tail bound~\eqref{EqnLedoux} for a constant $c \geq
 1$ large enough to ensure that $ \frac{c_2 \numproj}{8} \geq 2 \log
 M$. Taking the union bound over all $j \in [M]$ yields
\begin{align*}
\mprob[\opnorm{\Sketch U \Dbar^{1/2}} \geq 8 c \, \delcrit ] & \leq
c_1 e^{-\frac{c_2\numproj}{8} + \log M} \leq \; c_1 e^{-c'_2 \numproj}
\end{align*}
which completes the proof.

%%%%%%%%%%%%%%%%%%%%%%%%%%%%%%%

\paragraph{ROS case:}
Here we pursue a matrix Chernoff argument analogous to that in the
paper~\cite{tropp2011improved}. Letting $\radevec \in \{-1,
1\}^\numobs$ denote an i.i.d.\ sequence of Rademacher variables, the
ROS sketch can be written in the form $\Sketch = P H \diag(\radevec)$,
where $P$ is a partial identity matrix scaled by $\numobs/\numproj$,
and the matrix $H$ is orthonormal with elements bounded as $|H_{ij}|
\leq c/\sqrt{\numobs}$ for some constant $c$.  With this notation, we
can write
\begin{align*}
\opnorm{P H \diag(\radevec) \bar D ^{1/2}}^2 & = \opnorm{
  \frac{1}{\numproj} \sum_{i=1}^\numproj v_i v_i^T},
\end{align*}
where $v_i \in \real^\numobs$ are random vectors of the form
$\sqrt{\numobs} \Dbar^{1/2} \diag(\radevec) H e$, where $e \in
\real^\numobs$ is chosen uniformly at random from the standard
Euclidean basis.

We first show that the vectors $\{v_i\}_{i=1}^\numproj$ are uniformly
bounded with high probability.  Note that we certainly have $\max_{i
  \in [\numproj]} \|v_i\|_2 \leq \max_{j \in [\numobs]}
F_j(\radevec)$, where
\begin{align*}
F_j(\radevec) & \defn \sqrt{\numobs} \|\Dbar^{1/2} \diag(\radevec) H
e_j\|_2 \; = \; \sqrt{\numobs} \|\Dbar^{1/2} \diag(H e_j)
\radevec\|_2.
\end{align*}
Begining with the expectation, define the vector $\widetilde{\radevec}
= \diag(H e_j) \radevec$, and note that it has entries bounded in
absolute value by $c/\sqrt{\numobs}$. Thus we have,
\begin{align*}
\Exs[F_j(\radevec)] & \leq \Big[ \numobs \Exs[ \widetilde{\radevec}^T
    \Dbar \widetilde{\radevec} ] \Big]^{1/2} \; \leq \; c \,
\sqrt{\sum_{j = \effdim + 1}^\numobs \empkereig_j} \; \leq \; c
\sqrt{\numobs} \delcrit^2
\end{align*}
For any two vectors $\radevec, \radevec' \in \real^\numobs$, we have
\begin{align*}
\Big| F(\radevec) - F(\radevec') \Big| & \leq \sqrt{\numobs}
\|\radevec - \radevec'\|_2 \|\Dbar^{1/2} \diag(H e_j)\|_2 \; \leq
\delcrit.
\end{align*}
Consequently, by concentration results for convex Lipschitz functions of
Rademacher variables~\cite{Ledoux01}, we have
\begin{align*}
\mprob \Big[ F_j(\radevec) \geq c_0 \sqrt{\numobs} \delcrit^2 \log
  \numobs \Big] & \leq c_1 \CEXP{-c_2 \numobs \delcrit^2 \log^2
  \numobs}.
\end{align*}
Taking the union bound over all $\numobs$ rows, we see that
\begin{align*}
\max_{i \in [\numobs] } \|v_i\|_2 \; \leq \; \max_{j \in [\numobs]}
F_j(\radevec) & \leq 4 \sqrt{\numobs} \delcrit^2 \log(\numobs) 
\end{align*}
with probabablity at least $1 - c_1 \CEXP{-c_2 \numobs
    \delcrit^2 \log^2(\numobs)}$.
Finally, a simple calculation shows that $\opnorm{\Exs[v_1 v_1^T]}
\leq \delcrit^2$. Consequently, by standard matrix Chernoff
bounds~\cite{Tropp12,tropp2011improved}, we have
\begin{align} 
\label{EqnROSChernoff} 
\mprob \Big[ \opnorm{\frac{1}{\numproj} \sum_{i=1}^\numproj v_i v_i^T}
  \geq 2 \delcrit^2 \Big] & \leq c_1 \CEXP{- c_2 \frac{\numproj
    \delcrit^2}{\numobs \delcrit^4 \log^2(\numobs)}} + c_1 \CEXP{-c_2
  \numobs \delcrit^2 \log^2(\numobs)},
\end{align}
from which the claim follows.

%%%%%%%%%%%%%%%%%%%%%%%%%%%%%%%%%%%%%%%%%%%%%%%%%%%%%%%%%%%%%%%%%%%%%%%%%%

%% BIBLIOGRAPHY 
\bibliographystyle{abbrvnat}

\bibliography{clean_yun}
\end{document}